\pdfoutput=1

\documentclass[11pt]{article}

\usepackage[preprint]{acl}

\usepackage{times}
\usepackage{latexsym}
\usepackage{amssymb}

\usepackage{colortbl}

\usepackage{xcolor,pifont}
\newcommand*\colourcheck[1]{%
  \expandafter\newcommand\csname #1check\endcsname{\textcolor{#1}{\ding{52}}}%
}

\newcommand*\colourx[1]{%
  \expandafter\newcommand\csname #1check\endcsname{\textcolor{#1}{\ding{55}}}%
}

\colourx{red}

\colourcheck{blue}
\colourcheck{green}

\usepackage{microtype}

\usepackage{inconsolata}

\usepackage{graphicx}

\usepackage[utf8]{inputenc} 
\usepackage[T1]{fontenc}    
\usepackage{hyperref}       
\usepackage{url}            
\usepackage{booktabs}       
\usepackage{amsfonts}       
\usepackage{nicefrac}       
\usepackage{microtype}      
\usepackage{xcolor}         

\usepackage{graphicx}

\usepackage{amsmath}
\usepackage{bm}
\usepackage{graphicx} 
\usepackage{wrapfig} 
\usepackage{bbold}

\usepackage{tcolorbox}
\usepackage{multirow}
\usepackage{adjustbox}
\usepackage{subcaption}

\usepackage{float} 
\usepackage{booktabs}

\usepackage{tikz,lipsum}
\usepackage{listings}
\usepackage{subcaption}

\definecolor{citecolor}{HTML}{205a88}

\definecolor{cello}{HTML}{ffe6cc}

\makeatletter
\definecolor{customblue}{HTML}{a91d3a}

\definecolor{customred}{HTML}{d62728}
\newcommand{\red}[1]{\textcolor{customred}{#1}}
\definecolor{chart}{HTML}{206A9A}

\definecolor{arxiv}{HTML}{b31a1a}

\definecolor{sectioncolor}{HTML}{A91D3A}
\makeatother

\definecolor{OliveGreen}{rgb}{0.33, 0.42, 0.18}

\newtcolorbox{example2}[1][]{
  colback=arxiv!10!white,
  colframe=arxiv,
  floatplacement=floating,
  title=\centering #1
}

\newtcolorbox{example}[1][]{
  colback=chart!5!white,
  colframe=chart,
  floatplacement=floating,
  title=\centering #1,
}

\newtcolorbox{wronganswer}[1][]{
    colframe=arxiv,
    colback=arxiv!10!white,
    sharp corners,
    boxsep=0pt,
    left=5pt,
    right=5pt,
    top=6pt,
    bottom=6pt,
    boxrule=0pt,
    leftrule=4pt,
    #1
}

\newtcolorbox{correctanswer}[1][]{
    colframe=OliveGreen,
    colback=OliveGreen!10!white,
    sharp corners,
    boxsep=0pt,
    left=5pt,
    right=5pt,
    top=6pt,
    bottom=6pt,
    boxrule=0pt,
    leftrule=4pt,
    #1
}

\title{Chain of Functions: A Programmatic Pipeline for Fine-Grained Chart Reasoning Data}

\title{Chain of Functions: A Programmatic Pipeline for Fine-Grained Chart Reasoning Data}

\author{%
Zijian Li$^{12}$\thanks{Work done during an internship at Microsoft Research Asia.
}, Jingjing Fu$^{1}$, Lei Song$^{1}$, Jiang Bian$^{1}$,
\textbf{Jun Zhang}$^{2}$\thanks{Corresponding Author.}, \textbf{Rui Wang}$^{1\dag}$\\
$^1$Microsoft Research \quad
$^2$Hong Kong University of Science and Technology \\
\texttt{\{zijian.li\}@connect.ust.hk}, 
\texttt{\{jjfu,lesong,jiabia,ruiwa\}@microsoft.com}, \\
\texttt{eejzhang@ust.hk}
}

\begin{document}
\maketitle
\begin{abstract}

Visual reasoning is crucial for multimodal large language models (MLLMs) to address complex chart queries, yet high-quality rationale data remains scarce. Existing methods leveraged (M)LLMs for data generation, but direct prompting often yields limited precision and diversity.
In this paper, we propose \textit{Chain of Functions (CoF)}, a novel programmatic reasoning data generation pipeline that utilizes freely-explored reasoning paths as supervision to ensure data precision and diversity.
Specifically, it starts with human-free exploration among the atomic functions (e.g., maximum data and arithmetic operations) to generate diverse function chains, which are then translated into linguistic rationales and questions with only a moderate open-sourced LLM.
\textit{CoF} provides multiple benefits: 1) Precision: function-governed generation reduces hallucinations compared to freeform generation; 2) Diversity: enumerating function chains enables varied question taxonomies; 
3) Explainability: function chains serve as built-in rationales, allowing fine-grained evaluation beyond overall accuracy; 4) Practicality: eliminating reliance on extremely large models.
Employing \textit{CoF}, we construct the \textit{ChartCoF} dataset, with 1.4k complex reasoning Q\&A for fine-grained analysis and 50k Q\&A for reasoning enhancement.
The fine-grained evaluation on \textit{ChartCoF} reveals varying performance across question taxonomies for each MLLM, and the experiments also show that finetuning with \textit{ChartCoF} achieves state-of-the-art performance among same-scale MLLMs on widely used benchmarks.
Furthermore, the novel paradigm of function-governed rationale generation in \textit{CoF} could inspire broader applications beyond charts.
The code and data will be publicly available at \url{https://github.com/microsoft/Chain-of-Functions}.

\end{abstract}

\begin{figure}
    \centering
    \vspace{-0.1in}
    \includegraphics[width=0.98\linewidth]{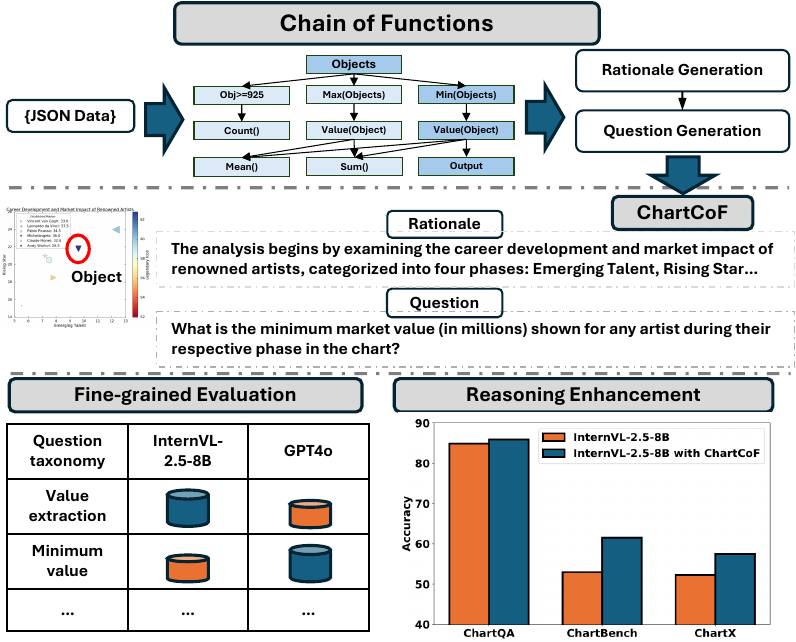}
    \vspace{-0.1in}
    \caption{Our proposed \textit{CoF} constructs a high-quality reasoning dataset \textit{ChartCoF} for the fine-grained evaluation and reasoning enhancement of MLLMs.}
    \label{fig:overview_chartcof}
    \vspace{-0.2in}
\end{figure}

\section{Introduction}
\label{intro}


Recent advancements in large language models (LLMs) \cite{chowdhery2023palm, dubey2024llama, guo2025deepseek} have paved the way for the development of multi-modal large language models (MLLMs) \cite{liu2024visual, bai2023qwenvl}, which have demonstrated a remarkable ability to understand visual semantics through the alignment between visual and embedding spaces. Despite this progress, current MLLMs exhibit limitations in their reasoning capabilities and
encounter difficulties in accurately interpreting charts in scholarly articles and financial documents \cite{xu2023chartbench, xia2024chartx}. This is particularly evident when they handle complex reasoning questions that necessitate accurate and step-by-step thought processes \cite{wangcharxiv}. 
The analysis in ChartQA \cite{masry2022chartqa}, as shown in Table \ref{eval_other_benchmark}, highlights a significant performance discrepancy between complex reasoning questions (Human set) and simpler perceptual questions (Augmented set). For instance, InternVL-2.5-8B \cite{chen2024internvl25} demonstrates a performance gap of nearly 20\%, which underscores the challenges that MLLMs face in bridging the gap between human-like reasoning and current computational capabilities.


Training with chain-of-thought (CoT) data has emerged as an effective strategy to enhance the reasoning abilities of MLLMs for chart understanding \cite{cot,zhang2024improve, zhangmultimodal_cot,REACHQA}. 
Nonetheless, high-caliber CoT data are scarce for chart reasoning, which require complete reasoning processes and accurate chart information (e.g., object values and positions) in the rationales \cite{dong2024insight, masry2024chartgemma, REACHQA}.
To generate CoT data, recent investigations have leveraged the capabilities of advanced (M)LLMs to autonomously produce questions, answers, and their corresponding rationales by either directly analyzing the charts or their textual descriptions with well-designed prompts \cite{liu2024mmc, REACHQA, masry2024chartgemma}. 
Despite these efforts, directly prompting (M)LLMs to generate questions and rationales based only on charts may result in low accuracy and limited diversity.
Moreover, relying on extremely large (M)LLMs poses a notable barrier to the data scalability.

In addition to the scarcity of CoT data for effective finetuning, the evaluation of MLLMs' reasoning capabilities remains underexplored. While current benchmarks have incorporated reasoning questions to evaluate the reasoning capabilities of MLLMs \cite{masry2022chartqa, xia2024chartx, wangcharxiv}, these questions often lack complexity and require only short reasoning chains to solve. 
More importantly, these benchmarks tend to gauge the reasoning performance in a broad sense with an overall accuracy metric, which overlooks the nuanced analysis of MLLMs' proficiency across questions that require varying reasoning chains. 
A fine-grained reasoning evaluation of the models' specific strengths and weaknesses on question taxonomies remains a valuable avenue for research.

In response to the scarcity of diverse and high-caliber reasoning datasets for the fine-grained evaluation and enhancement of chart reasoning, as presented in Fig. \ref{fig:overview_chartcof}, we introduce a novel automatic reasoning data synthesis pipeline named \textit{Chain of Functions (CoF)}. 
Unlike prior methods that rely on end-to-end LLM prompting, our approach first systematically explores chart elements through a set of atomic functions to ensure correct and diverse reasoning paths and then translate them into language, which greatly reduces hallucinations and enables more precise supervision.
Concretely, \textit{CoF} encompasses two key processes: \textit{program-based functional discovery} and \textit{reverse linguistic CoT data synthesis}.
In program-based functional discovery, we carefully design atomic functions and their corresponding conditions, which are intelligently combined to form a coherent function chain based on a chart.
Then in the reverse linguistic CoT data synthesis process, these function chains are translated into natural language instructions using LLMs in a reverse manner, with rationales first, and then questions. 
This method ensures the precision of questions, rationales, and answers.
Crucially, since the reasoning process is determined by the function chain rather than by generative prompts alone, we can leverage a moderate open-sourced LLM (Qwen2.5-32B-instruct \cite{yang2024qwen25} used in experiments) for the linguistic transfer, greatly lowering dependence on extremely large models.
Furthermore, \textit{CoF} effectively bridges structured reasoning and language modeling, with potential applications beyond charts.



\textbf{Key contributions:} 
1) Our proposed reasoning data generation pipeline \textit{CoF} greatly ensures explainability, precision, and diversity of generated reasoning data, thus enabling the fine-grained evaluation and reasoning enhancement for MLLMs.

2) We introduce \textit{ChartCoF}, which encompasses an extensive variety of over 19 chart types, with a test set comprising 648 charts paired with 1,451 Q\&A pairs and a training set featuring 18,349 charts with 50,329 Q\&A pairs for fine-grained evaluation and model finetuning.


3) The fine-grained evaluation reveals the weak performance of existing MLLMs on the complex reasoning questions of \textit{ChartCoF} and provides deep insight into their skilled and unskilled question taxonomies. 

4) Extensive experiments demonstrate that after finetuning with \textit{ChartCoF}, MLLMs achieve state-of-the-art performance among same-scale MLLMs in three widely used benchmarks on charts.
Additional analysis reveals the significant improvement of finetuned MLLMs with \textit{ChartCoF} on out-of-distribution (OOD), model scalability, and data scalability evaluations.


\section{Related Works}

\textbf{Multimodal large language models (MLLMs)} have aligned the vision space with the embedding space of LLMs for visual understanding \cite{vaswani2017attention, radford2018improving, brown2020language, zhang2022opt, chowdhery2023palm, dubey2024llama, team2023internlm, bai2023qwen, yin2023survey}, which is normally achieved via connectors, e.g., Q-Former \cite{li2023blip} or MLP \cite{bai2023qwenvl}.
With connectors, Mini-GPT4, mPLUG-Owl, and InstructBLIP have extended language-only instruction tuning to multimodal tasks. 
LLaVA \cite{liu2024visual,li2024llavaonevision} also maps visual features into the LLaMA \cite{touvron2023llama} embedding space using a linear layer.
Its modularization and high efficiency in training make it a popular architecture of MLLMs.
Despite the impressive achievements of existing open-sourced MLLMs, e.g., QwenVL \cite{bai2023qwenvl, wang2024qwen2vl}, InternVL \cite{chen2024internvl,chen2024internvl25}, and DeepSeek-VL \cite{lu2024deepseekvl, wu2024deepseekvl2} in common multimodal tasks like VQA \cite{antol2015vqa} and image captioning \cite{vinyals2015show}, they focus more on perception tasks while paying less attention to the visual reasoning capabilities, especially for chart understanding.
In this work, we focus on improving and evaluating the reasoning capabilities for MLLMs on charts.






\textbf{Chart reasoning}
refers to dealing with intricate tasks related to both chart-related and commonsense knowledge \cite{xu2024chartmoe, REACHQA}.
The early two-stage inference studies first extracted structural information like tables and markdowns and then leveraged textual information for downstream understanding \cite{liu2023matcha, liu2023deplot, lee2023pix2struct, wang2023domino}.
Afterwards, the unified MLLMs, e.g., OneChart \cite{chen2024onechart}, UniChart \cite{masry2023unichart}, ChartMoE \cite{xu2024chartmoe}, and TinyChart \cite{zhang2024tinychart}, are trained to handle varying chart-related tasks.
However, these methods focus on the perception capabilities of MLLMs and overlook the reasoning capabilities.
In this work, we aim to improve and evaluate the reasoning capabilities from a data aspect by generating high-quality CoT data.
Many studies have utilized powerful proprietary GPT or Gemini series to generate reasoning instruction tuning data \cite{xu2023chartbench, liu2024mmc, xia2024chartx, han2023chartllama, masry2024chartinstruct, masry2024chartgemma, CHOPINLLM, REACHQA, liu2024chartthinker, shen2024rethinking}. 
However, directly prompting (M)LLMs based only on charts may affect the precision and diversity of training data.
The excessive reliance on extremely large models also poses a significant barrier to data generation.
To generate accurate Q\&A, many methods attempted to manually set up templates to obtain Q\&A in an end-to-end manner \cite{huang2024evochart, methani2020plotqa, meng2024chartassisstant, LAMENDA}.
Nevertheless, the predefined question templates follow a fixed pattern and may lead to limited diversity, affecting the generalization of MLLMs.
In contrast, we propose a functional discovery workflow to ensure the diversity of reasoning paths and a reverse linguistic CoT data synthesis to enhance the reality and diversity of generated questions.
The extra supervision of function chains during generation also refrains from the reliance on extremely large (M)LLMs.
A more detailed comparison between \textit{ChartCoF} and existing datasets is presented in Appendix \ref{appendix_dataset_comparison}.




















\section{Chain of Functions}

\begin{figure*}
\vspace{-0.2in}
    \centering
    \includegraphics[width=0.95\textwidth]{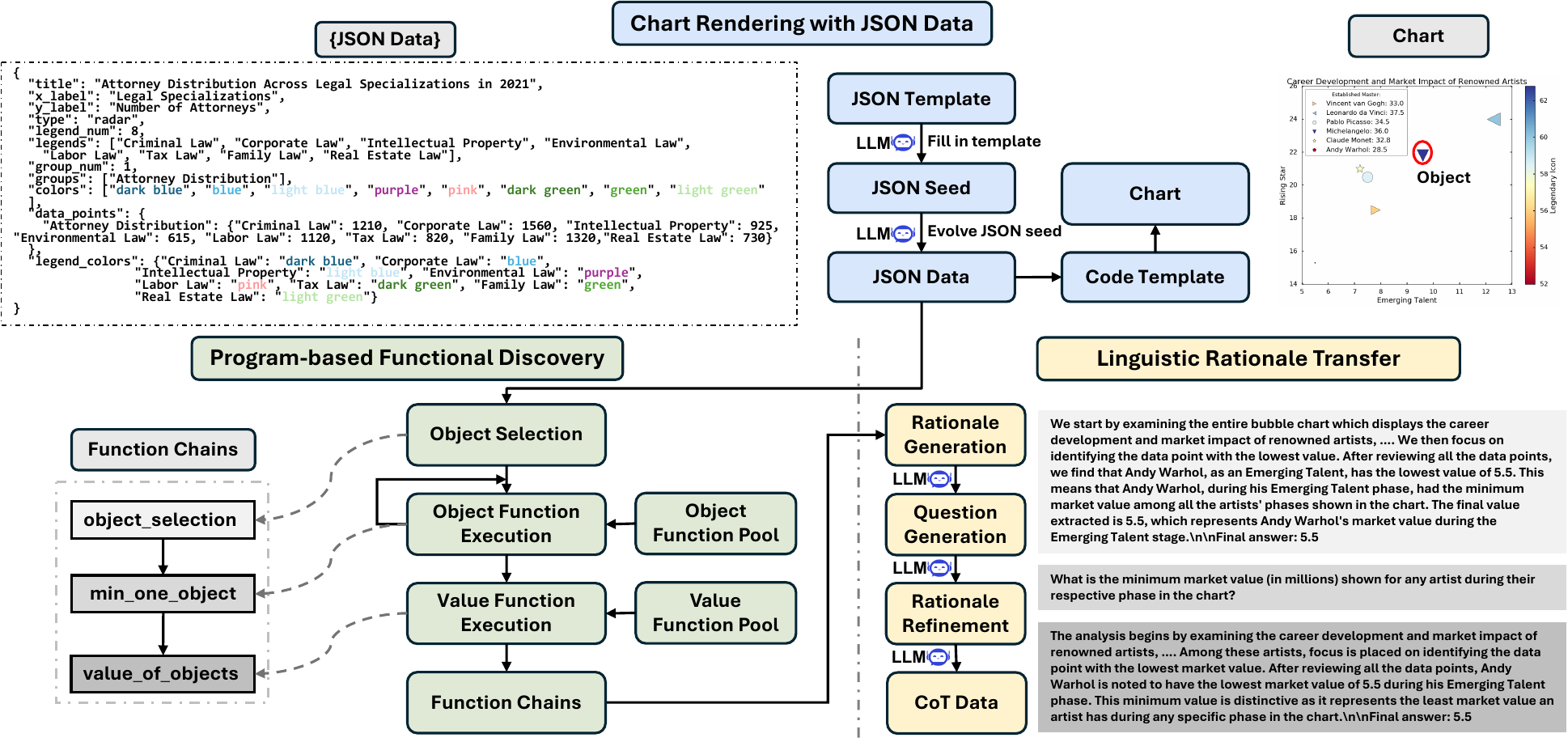}
    \vspace{-0.1in}
    \caption{Overview of \textit{chain of functions}. We prompt LLMs to fill in the JSON template to construct JSON seed and evolve (modify) it to more accurate and diverse JSON data. The JSON data are then used to generate function chains through functional discovery. The function chains are then transferred to CoT data by prompting LLMs.}
    \label{fig:cof}
    \vspace{-0.2in}
\end{figure*}

In this section, we propose the reasoning data synthesis pipeline \textit{chain of functions (CoF)}, including chart rendering, program-based function discovery, and reverse linguistic CoT data synthesis.
The overview of \textit{CoF} is presented in Fig. \ref{fig:cof}.



\subsection{Chart Rendering with JSON Data}
To ensure the consistency between charts and generated CoT data, we leverage JSON data as the intermediate representation, which is then used for chart rendering and reasoning data generation. 

\textbf{JSON template.}
We predefine the essential elements of charts in a structural presentation for subsequent chart rendering and CoT data generation, which includes title, $x$ label, $y$ label, chart type, legend number, legend list, group number, group list, data points, colors, and legend colors.
For some special charts, e.g., boxes, candlesticks, and node links, we include the additional elements. 
The JSON templates for all chart types are displayed in Appendix \ref{json_template}.
The elements of the chart provide ground-truth information for chart rendering and subsequent reasoning data generation.

\textbf{JSON generation.}
To generate realistic information for charts, the titles are generated using LLMs for each chart type. 
These titles are then used to generate the JSON files by prompting LLMs.
To ensure the diversity of JSON data, we randomly sample group number, legend number, and colors for JSON templates and prompt LLMs to only fill in the rest of the elements that require realistic knowledge, e.g., group list, legend list, and data points, producing JSON seed files.
To scale up, we further prompt LLMs to evolve the JSON seed and generate more realistic and accurate JSON data.
All the prompts for JSON seed generation and JSON evolvement are presented in Appendix \ref{prompt_json_generation_evolement}.

\textbf{Chart rendering with code templates.}
To avoid visual conflicts of chart images, we carefully design the code template for each type of chart.
With the detailed information in JSON data and the well-designed code templates, we generate the chart image for each JSON file.
To ensure the diversity of chart images, we use varying code libraries for chart rendering, including matplotlib, sklearn, mplfinance, plotly, seaborn, and networkx.
For each type of chart, we set up different plotting styles, color transparency, and element locations.
The chart examples are shown in Appendix \ref{examples}.



\begin{table*}[t]
\vspace{-0.3in}
\resizebox{\textwidth}{!}{
\begin{tabular}{cccccccccccccccc}
\toprule
\multirow{2}{*}{Data split}  & \multirow{2}{*}{\begin{tabular}[c]{@{}c@{}}\#chart \\types\end{tabular}} & \multirow{2}{*}{\#charts} & \multirow{2}{*}{\#Q\&A} & \multirow{2}{*}{\begin{tabular}[c]{@{}c@{}}\#words of \\ rationales\end{tabular}} & \multicolumn{6}{c}{Lengths of function chains}                        & \multirow{2}{*}{\begin{tabular}[c]{@{}c@{}}\#Function  \\ chains\end{tabular}} & \multirow{2}{*}{\#Functions}       & \multicolumn{3}{c}{Question type} \\ \cmidrule(lr){6-11} \cmidrule(lr){14-16}
         &       &   & &         & 2       & 3       & 4      & 5       & 6       & $\geq$7 &                            &        & Binary   & NQA      & Text    \\ \toprule
Training set  & 19   & 18,349 & 50,329  & 66.62         & 38.58\% & 26.42\% & 2.0\%  & 20.87\% & 9.83\%  & 2.30\%           & 3,134        & 107                       & 16.68\%  & 55.41\%  & 27.91\% \\
Test set   & 19   & 648   &   1,451       & -        & 22.54\% & 20.74\% & 9.72\% & 16.68\% & 13.58\% & 16.75\%          & 728       & 107                         & 16.40\%  & 67.88\%  & 15.72\% \\ \toprule
\end{tabular}}
\vspace{-0.15in}
\caption{Statistics of training and test sets. \textit{CoF} enables synthetic data with long and diverse reasoning paths. Detailed statistics of each chart type and function taxonomy are present in Appendix \ref{data_splitting} and Appendix \ref{appendix_function}, respectively.}
\vspace{-0.2in}
\label{stat}
\end{table*}


\subsection{Program-based Functional Discovery}
\textit{CoF} conducts human-free exploration in the function pools, resulting in diverse and numerous function chains. 
This mirrors the pattern of chart understanding of humans: we select partial or all of the objects from the chart, extract information from them, recognize their trend and pattern, or conduct comparison and calculation between them.
The function chain is discovered via a three-step workflow, which is elaborated as below.

1) \textbf{Object selection.} We regard each data point in the charts as one object. At the beginning of the workflow, partial or all of the objects from the chart are selected using chart information, including groups, legends, and colors. 
The selected objects are used for sequential function execution.

2) \textbf{Object function execution.} We define the functions with objects as input as the object functions to imitate reasoning processes on charts, e.g., an information extraction process to get the legend of a data point and a pattern recognition process to get the maximum data value. 
The selected objects are greedily input into the object functions and obtain the corresponding output, which results in a functional triplet $<input, function, output>$.
The $input$ are objects, and the $output$ can be objects, numerical values, strings, or booleans.
The executed functional triplet is recorded and spliced with the previously executed function triplet to form a function chain.
The function chain with the final output of numerical values, strings, or booleans is regarded as a complete chain.
Notably, the function chain with object output continues to perform step 2) and executes object functions again, enabling a longer function chain.

3) \textbf{Value function execution.}
We define the functions with numerical values as input as the value functions to imitate the reasoning processes of value comparison and arithmetical operation.
The function chains with the output of numerical values after step 2) are greedily input into the value functions from the value function pool and obtain the final answer.
We also allow multiple separate function chains to execute value functions jointly to achieve the combination of them, resulting in a longer and more complex function chain.

To ensure the realisticity of function chains, we set up the execution conditions for each function and explore feasible function chains that meet these conditions.
The details of functions for object selection, object functions, and value functions are present in Appendix \ref{appendix_function}.
With the above three-step workflow, we generate accurate and diverse function chains, which also provide explainability for the subsequent CoT data generation.

\subsection{Reverse Linguistic CoT Data Synthesis}
To generate precise and realistic rationales and questions, we transfer function chains to linguistic CoT data in a reverse manner by first rationales, then questions, and finally refining rationales.

1) \textbf{Linguistic rationale transfer.} We prompt LLMs to transfer function chains to linguistic rationales. To make LLMs better understand each function and generate more precise lingustic rationales, we also include the description of each function into the prompt.

2) \textbf{Question generation.} We prompt LLMs to generate realistic questions using JSON data, function chains, and the generated rationales.
The chart information and the generated lingustic rationales enable LLMs to better understand the reasoning process and generate more precise questions.

3) \textbf{Rationale refinement.} We empirically found that initial-generated rationales are still function-like and redundant. Thus, we prompt LLMs to concisely refine the initial-generated rationales based on function chains and questions, making them align better with MLLMs.
The effectiveness of rationale refinement is discussed in Appendix \ref{ablation_rationale_refinement}.

Under the supervision of function chains, the reverse linguistic CoT data synthesis can be regarded as a translator task between function chains and linguistic CoT data, without the requirement of extremely large models.
All the prompts for CoT data synthesis are present in Appendix \ref{prompt_json_generation_evolement}.


\begin{table*}[t]
\vspace{-0.3in}
\centering
\resizebox{0.95\textwidth}{!}{
\begin{tabular}{ccccccccc}
 \toprule[2.0pt]
\multirow{2}{*}{Models}        & \multicolumn{2}{c}{Annotation} & \multicolumn{3}{c}{Task} & \multicolumn{2}{c}{Chart type}  & \multirow{2}{*}{Avg.} \\ \cmidrule(lr){2-3} \cmidrule(lr){4-6} \cmidrule(lr){7-8}
                                  & w.o.           & w.            & Binary  & NQA    & Text  & Regular         & Extra   &           \\  \midrule  
\multicolumn{9}{>{\columncolor{gray!20}}c}{proprietary models}                                                                                                                         \\ \midrule 
GPT4o \cite{achiam2023gpt}                                                & 42.16          & \underline{76.85}         & \underline{81.51}   & \underline{55.74}  & \underline{57.46} & \underline{65.17}           & \underline{54.48}      & \underline{60.23}       \\
GPT4V \cite{achiam2023gpt}                                                  & 26.62         & 59.26         & 68.49   & 39.59  & 35.09 & 46.86           & 39.85   & 43.63        \\
 \midrule 
\multicolumn{9}{>{\columncolor{gray!20}}c}{Open-sourced models}                                                                                                                         \\ \midrule 
InternLM-XComposer-2.5-7B \cite{zhang2024internlm_xcomposer_25}                                 & 34.67          & 51.72         & 61.34   & 42.34  & 30.26 & 50.19           & 35.82     & 43.56      \\
DeepSeek-VL2-small \cite{wu2024deepseekvl2}                           & 18.41          & 24.87         & 55.04   & 12.18  & 28.51 & 20.49           & 23.28          & 21.78           \\
LLaVA-v1.6-mistral-7B \cite{liu2024visual}                                & 22.73          & 30.16         & 50.84   & 21.92  & 21.49 & 27.53           & 25.53       & 26.60       \\
Qwen2VL-7B \cite{wang2024qwen2vl}                                     & 39.28          & 58.99         & 78.15   & 44.77  & 40.35 & 55.19           & 42.99            & 49.55       \\
InternVL-2.5-8B \cite{chen2024internvl25}                                  & 36.98          & 63.23         & 69.33   & 48.63  & 39.91 & 59.80           & 40.00       & 50.65         \\
CogVLM2-7B \cite{hong2024cogvlm2}                                 & 25.47          & 46.43         & 65.97   & 32.18  & 23.68 & 37.90           & 34.63               & 36.39        \\ \midrule
\multicolumn{9}{>{\columncolor{gray!20}}c}{Chart-specific models}                                                                                                                    \\ \midrule 
ChartInstruct-7B \cite{masry2024chartinstruct}                                         & 13.52          & 16.01         & 55.88   & 7.92   & 1.75  & 13.96           & 15.82       & 14.82    \\
ChartVLM-14.3B \cite{xia2024chartx}                                     & 20.29          & 23.15         & 49.16   & 18.48  & 7.46  & 24.07           & 19.10             & 21.78       \\
ChartGemma-2B \cite{masry2024chartgemma}                                            & 25.04          & 35.85         & 58.40   & 26.90  & 17.98 & 35.08           & 25.52  & 30.67         \\
ChartMoE-8B \cite{xu2024chartmoe}                            & 34.96          & 50.00         & 72.27   & 38.17  & 32.02 & 47.50           & 37.31    & 42.80        \\
InternVL-2.5-8B + \textbf{ChartCoF}                             & \textbf{63.74}          & \textbf{79.50}         & \textbf{89.50}   & \textbf{68.63}  & \textbf{67.98} & \textbf{77.85}           & \textbf{65.07} & \textbf{71.95}  \\ \toprule[2.0pt]    
\end{tabular}}
\vspace{-0.15in}
\caption{Accuracy performance of MLLMs with CoT prompts on \textbf{ChartCoF}. The best and second-best scores are highlighted in \textbf{bold} and \underline{underline}, respectively.}
\vspace{-0.2in}
\label{eval_chartcof}
\end{table*}

\section{ChartCoF}
Employing \textit{CoF}, we construct a dataset named \textit{ChartCoF}, which encompasses an extensive variety of 19 chart types, with a test set comprising 648 charts and 1,451 Q\&As and a training set featuring 18,349 charts and 50,329 Q\&As.
We adopt Qwen2.5-32B-instruct \cite{yang2024qwen25} for data generation in \textit{CoF}.
The statistics of \textit{ChartCoF} from the aspects of charts, function chains, and questions are described in Table \ref{stat}.

\textbf{Chart types.} \textit{ChartCoF} covers all the chart types that can be represented using the JSON format, with totally 19 chart types. We categorize the chart types into two groups based on their usage frequency. \textbf{Regular chart types:} We include bar charts (with single and multiple groups of bars and stacked bars), line charts (with single and multiple lines), and pie charts.
These six chart types are commonly used in most of the existing datasets \cite{masry2022chartqa, methani2020plotqa}.
\textbf{Extra chart types:} We also cover the complex chart types on existing datasets \cite{xu2023chartbench, xia2024chartx}, including rings, radar, rose, candlestick, 3D-bar, treemap, funnel, heatmap, treemap, box, area, bubble, multi-axes, and node link. Note that each chart type can be \textbf{annotated or not} if allowed.

\textbf{Question types.}
\textit{ChartCoF} focuses on MLLMs' reasoning capabilities and thus adopts chart-related question answering (QA) tasks. 
We categorize the question types based on the contexts of output.
\textbf{Binary:} Binary questions aim to assess the correctness of arguments.
\textbf{Text:} For text questions, the answers are from the elements of charts, such as group names and legends.
\textbf{Numerical question answering (NQA):} We also provide numerical questions that contain numerical computing processes.

\textbf{Function chains:} In \textit{ChartCoF}, 99 object functions and 8 value functions are used to construct function chains, which results in 3,134 and 728 function chains for the training set and test set, respectively. 
The length of these function chains ranges from 2 to 13, constructing the rationales with 66.62 average words for the training set.

\textbf{Evaluation metrics:} 
\label{Eval}
We follow ChartQA \cite{masry2022chartqa} and ChartX \cite{xia2024chartx} to adopt accuracy (Acc) as the evaluation metric and allow 5\% margin for numerical responses.
For those MLLMs with weak instruction-following capabilities that cannot output the final answer in a correct format, we additionally prompt GPT4o to extract the final answer \cite{xu2023chartbench}. 
This makes the 5\% margin feasible for these MLLMs to ensure a fair comparison.
The prompt for answer extraction is presented in Appendix \ref{prompt_answer_extraction}.

\section{Experiments}
Employing \textit{ChartCoF}, we provide fine-grained evaluation on varying question taxonomies for existing MLLMs.
The evaluation of other benchmarks, out-of-distribution (OOD) analysis, and model and data scalability are leveraged to demonstrate the effectiveness of \textit{ChartCoF} in enhancing reasoning capabilities.

\subsection{Experimental Setups}


\textbf{Benchmarks.}
Besides our proposed \textit{ChartCoF}, we also evaluate the MLLMs in existing benchmarks about chart reasoning, including ChartQA \cite{masry2022chartqa}, ChartBench \cite{xu2023chartbench}, and ChartX \cite{xia2024chartx}.
For ChartQA and ChartBench, we adopt all the test samples. For ChartX, we select only the QA task samples  for evaluation and leave other unrelated tasks like chart redrawing.
By following the evaluation metrics of these benchmarks, we allow 5\% margin for the NQA tasks, and \textit{Acc+} is used to evaluate the binary tasks in ChartBench \cite{xu2023chartbench}.
Since we find that inferencing with a CoT strategy cannot improve performance for baseline MLLMs, we prompt them to direct output final answers on these three benchmarks by following the recent work \cite{xu2024chartmoe}.
Since the questions in the Augmented set of ChartQA are  the perceptual questions without the need of thinking, we prompt our finetuned MLLMs to direct output the answer.

\textbf{Models and baselines.}
We evaluate a wide range of MLLMs in \textit{ChartCoF} and other benchmarks across three categories:
1) \textbf{Proprietary models}, including GPT4o \cite{achiam2023gpt} and GPT4V \cite{achiam2023gpt};
2) \textbf{Open-sourced MLLMs}, including InternLM-XComposer-2.5 \cite{zhang2024internlm_xcomposer_25}, DeepSeek-VL2-small \cite{wu2024deepseekvl2}, LLaVA-v1.6-mistral-7B \cite{li2024llavaonevision}, CogVLM2 \cite{hong2024cogvlm2}, Qwen2VL-7B \cite{wang2024qwen2vl}, and InternVL-2.5-8B \cite{chen2024internvl25};
3) \textbf{Chart-specific MLLMs}. including ChartInstruct \cite{masry2024chartinstruct}, ChartVLM \cite{xia2024chartx}, ChartGemma \cite{masry2024chartgemma}, ChartMoE \cite{xu2024chartmoe}, TinyChart \cite{zhang2024tinychart}, ChartLlama \cite{han2023chartllama}, and ChartAst \cite{meng2024chartassisstant}.

\textbf{Experiment details.} To demonstrate the effectiveness of \textit{ChartCoF} in enhancing reasoning capabilities of MLLMs, we finetune two off-the-shelf MLLMs, i.e., InternVL-2.5-8B and Qwen2VL-7B, with the training set of \textit{ChartCoF}.
We finetune them in one epoch by tuning the LLM part and freezing the vision encoder and projector in 4 A100-80G GPUs, with a batch size of 32, a learning rate of $5e-6$, and a weight decay of $0.01$.
To achieve better instruction-following capabilities, we adopt a CoT prompt ``\textit{Think step by step to generate the rationales, and then answer the question using a single word or phrase. The output format is Rationale: [Your Rationale] Answer: [Your Answer]}'' for both finetuning and inference.
We also leverage self-consistency technologies to further enhance the performance by setting a temperature of 0.8 and selecting the final answer with a majority vote of 5 attempts.
The evaluation metrics on \textit{ChartCoF} can be referred to in Section \ref{Eval}.

\begin{figure*}
\vspace{-0.5in}
	\begin{subfigure}[b]{0.32\textwidth}
	\centering
	\includegraphics[width =\textwidth]{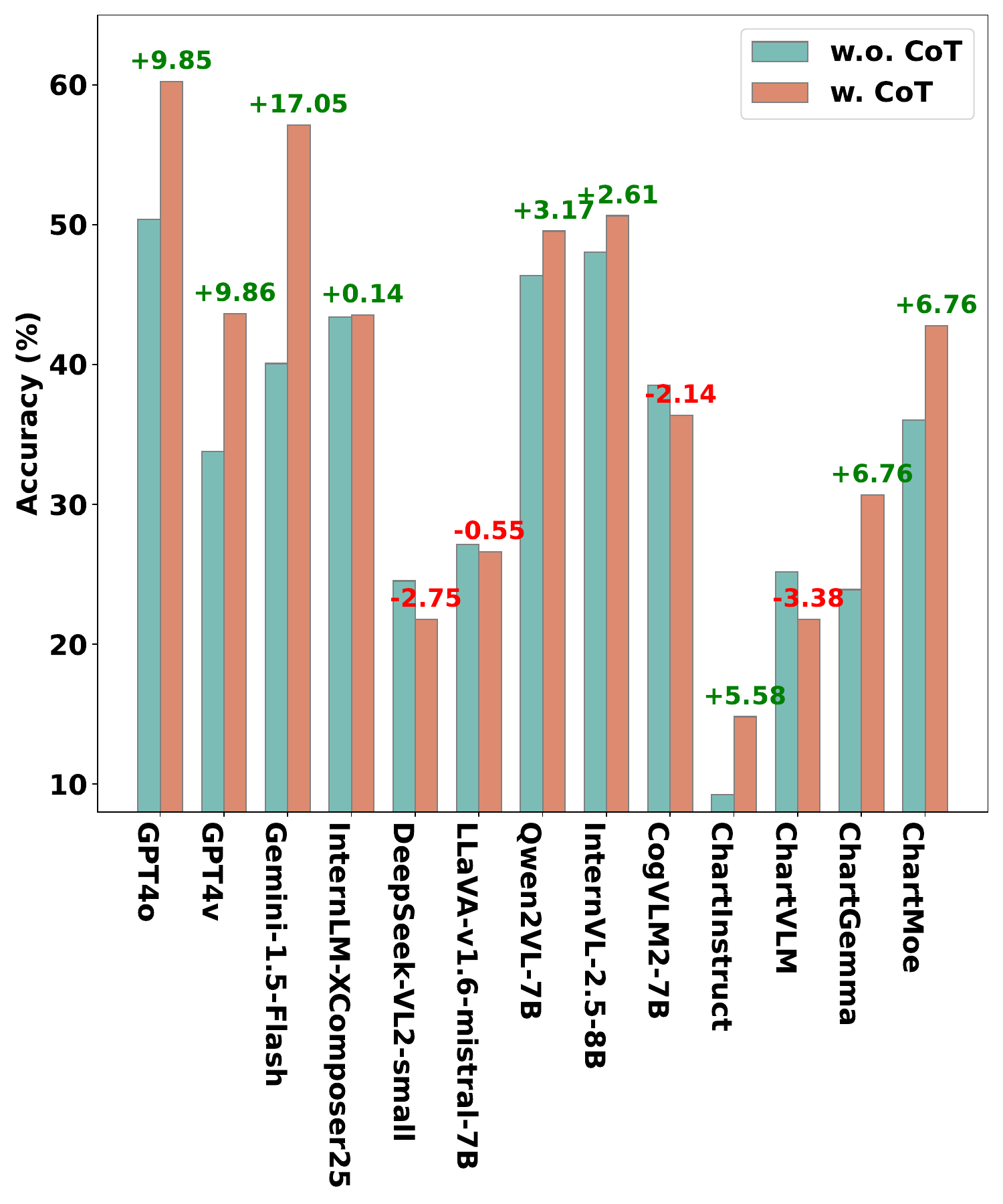}
    \vspace{-0.3in}
	\caption{}
	\label{fig:cot}
	\end{subfigure}
	\begin{subfigure}[b]{0.32\textwidth}
	\centering
	\includegraphics[width =\textwidth]{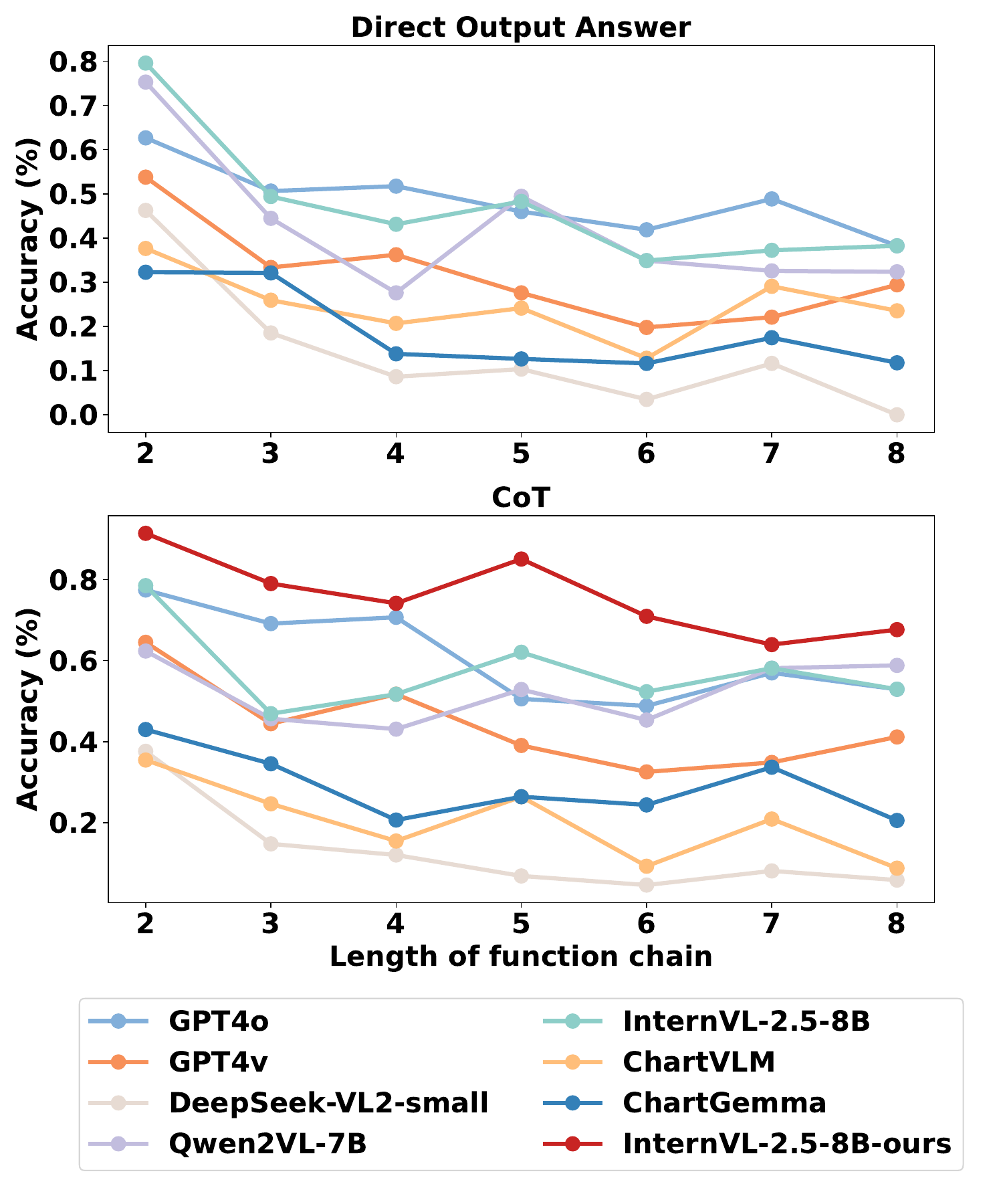}
    \vspace{-0.3in}
	\caption{}
	\label{fig:step_num}
	\end{subfigure}
	\begin{subfigure}[b]{0.32\textwidth}
	\centering
	\begin{subfigure}[b]{\textwidth}
	\centering
	\includegraphics[width = \textwidth]{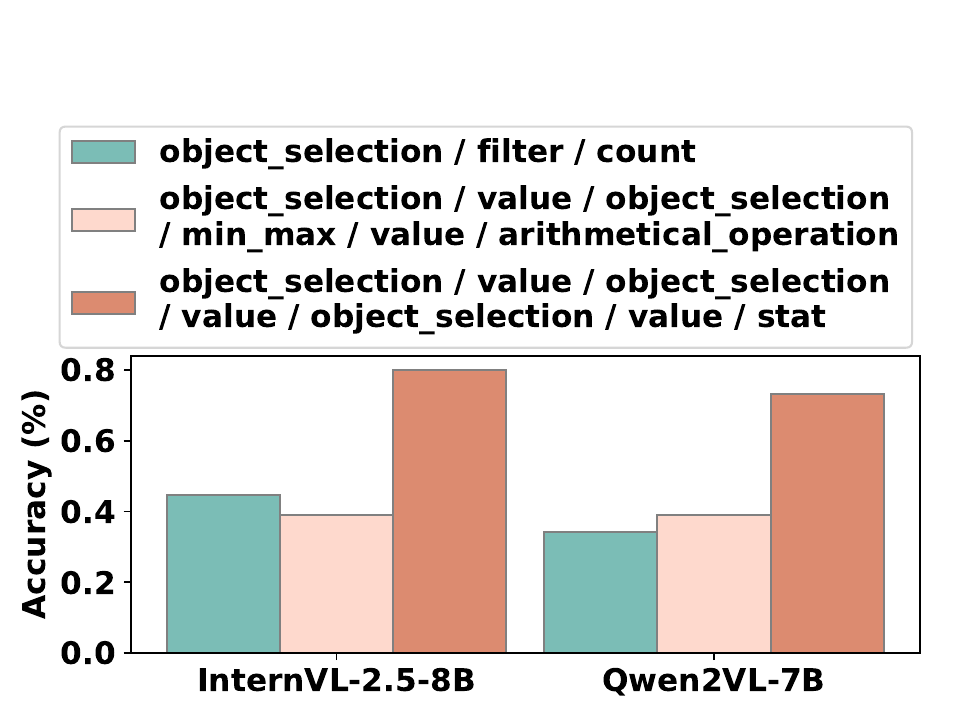}
    \vspace{-0.3in}
	\caption{}
	\label{fig:acc_function_chain}
	\end{subfigure}
    
    \begin{subfigure}[b]{\textwidth}
	\centering
	\includegraphics[width = \textwidth]{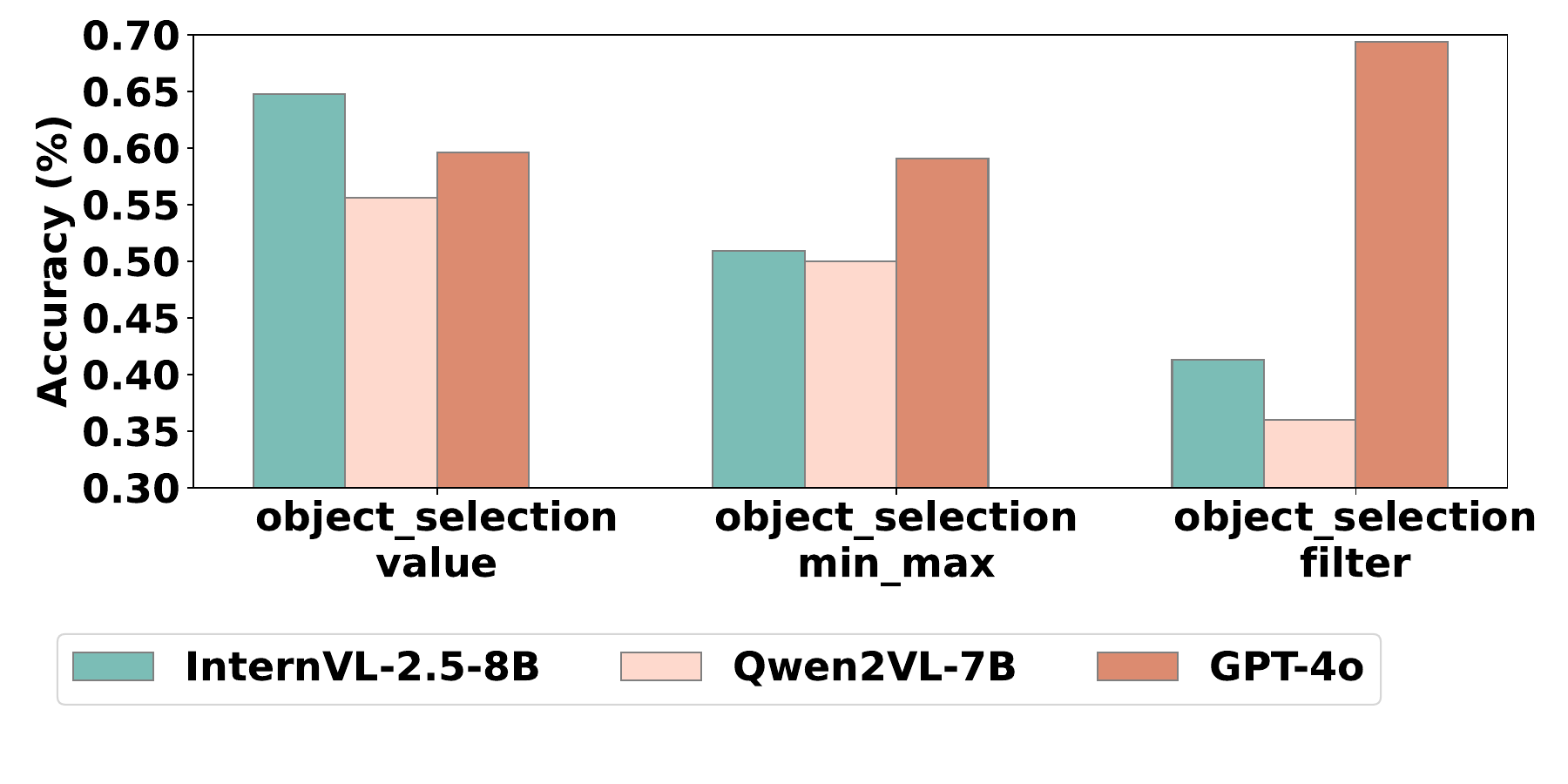}
    \vspace{-0.2in}
	\caption{}
	\label{fig:function combination}
	\end{subfigure}
	\end{subfigure}
\vspace{-0.1in}
\caption{(a) Accuracy of MLLMs with and without CoT strategies on ChartCoF. (b) Accuracy of MLLMs across questions with different lengths of function chains. (c) Accuracy of MLLMs across questions with different function chains. Some corresponding examples are presented in Table \ref{example_for_function_chain}. (d) Accuracy of MLLMs across questions with different short function chains.}
\label{fig:label}
\end{figure*}


\begin{table*}[]
\vspace{-0.1in}
\resizebox{\textwidth}{!}{
\begin{tabular}{|l|l|l}
\cline{1-2}
Function chain                                                                             & Examples                                                                                                                                                                                                                                                 &  \\ \cline{1-2}
object\_selection / filter / count\_of\_objects                 & \begin{tabular}[c]{@{}l@{}}How many legal specializations have more than 925 attorneys according to the 'Attorney Distribution' \\ group in the chart? (see Example \ref{example 1})\end{tabular}                                                                                      &  \\ \cline{1-2}
object\_selection / value / object\_selection / min\_max / value / arithmetical\_operation & \begin{tabular}[c]{@{}l@{}}What is the ratio of Social Media traffic on Sunday compared to the day with the second highest Social\\ Media traffic? (see Example \ref{example 2})\end{tabular}                                                                                          &  \\ \cline{1-2}
object\_selection / value / object\_selection / value / object\_selection / value / stat    & \begin{tabular}[c]{@{}l@{}}If we find the number of visitors using 'Other' devices in April, the number of 'Mobile' visitors in March, \\ and the number of  'Tablet' visitors in May,  what is the median value among these three numbers? (see Example \ref{example 3})\end{tabular} &  \\ \cline{1-2}
\end{tabular}}
\vspace{-0.1in}
\caption{Examples of the function chains in Fig. 
\vspace{-0.1in}
\ref{fig:acc_function_chain}.}
\label{example_for_function_chain}
\end{table*}


\subsection{Fine-grained Evaluation on ChartCoF}

\textbf{Main results on ChartCoF.}
We evaluate MLLMs with and without CoT strategies and present the results in Fig. \ref{fig:cot}.
Results show that most of the MLLMs improve the accuracy performance with CoT.
However, the existing MLLMs, including the proprietary and chart-specific models, still struggle with the complex reasoning questions on \textit{ChartCoF}.
As the results in Table \ref{eval_chartcof}, all of the MLLMs with CoT achieve low accuracy.
Among them, GPT4o achieves the best performance, with an accuracy of 60.23\%, a testament to its significant reasoning capabilities.
We also observe that the chart-specific models achieve lower accuracy compared to other models, demonstrating the necessity of our proposed \textit{ChartCoF} for reasoning enhancement on these complex reasoning questions.
After finetuning InternVL-2.5-8B with \textit{ChartCoF}, it achieves the state-of-the-art performance.

\begin{table*}[]
\resizebox{\textwidth}{!}{
\vspace{-0.2in}
\begin{tabular}{cccccccc}
 \toprule[2.0pt]
\multirow{2}{*}{{Models}}                                            & \multicolumn{3}{c}{{ChartBench}}  & \multicolumn{3}{c}{{ChartQA}}    & {ChartX} \\
\cmidrule(lr){2-4}  \cmidrule(lr){5-7} \cmidrule(lr){8-8} 
      & Reg.   & Extra          & Avg.           & Human          & Aug.  & Avg.           & NQA             \\ \midrule 
GPT4o \cite{achiam2023gpt}                                                                    & 60.02  & \textbf{58.89}          & 59.45          & -              & -     & 84.70           & 46.60  \\
ChartVLM-14.3B \cite{xia2024chartx}                                                             & 15.16  & 8.38           & 11.96          & 42.08          & 82.48 & 62.28          & 40.71  \\
ChartLlama-13B \cite{han2023chartllama}                                                           & 20.99  & 21.71          & 21.31          & 58.40          & 93.12 & 75.76          & 13.80  \\
ChartGemma-3B \cite{masry2024chartgemma}                                                             & 39.89  & 42.27          & 38.46          & 67.84          & 85.28 & 76.56          & 35.15           \\
TinyChart-3B \cite{zhang2024tinychart}                                                              & 26.71  & 22.56          & 22.51          & 70.24          & 91.04 & 76.80           &        40.10         \\
ChartAst-13B \cite{meng2024chartassisstant}                                                              & 3.82     & 1.58            & 2.81             & 64.88          & 93.12 & 79.00           & 30.99  \\
ChartMoE-8B \cite{xu2024chartmoe}                                                                   & 56.31  & 55.58          & 51.67          & 78.32          & 90.96 & 84.64          & 46.62  \\ \midrule  \midrule 
InternVL-2.5-8B \cite{chen2024internvl25}                                                        & 62.23 & 41.73          & 52.96          & 75.20           & \textbf{94.56} & 84.88          & 52.26           \\
InternVL-2.5-8B + \textbf{ChartCoF}                                                     & 67.41  & 54.63          & 61.63 \color{blue}{(+8.67) }        & \underline{77.12} & \underline{94.48} & \underline{85.80} \color{blue}{(+0.92) } & 58.77 \color{blue}{(+6.51)}           \\
+self-consistency  & \textbf{70.42}  & 56.97          & \textbf{64.33} \color{blue}{(+11.37) }          & \textbf{79.04}          & 94.32  & \textbf{86.68} \color{blue}{(+1.80) }         & 58.77 \color{blue}{(+6.51)}          \\ \hline
Qwen2VL-7B \cite{wang2024qwen2vl}                                                          & 63.13  & {56.23} & 60.01          & 73.28          & 94.40  & 83.84          & 52.17           \\
Qwen2VL-7B + \textbf{ChartCoF}                                                              & 67.01  & 55.35          & {61.73} \color{blue}{(+1.72) } & 76.00           & 93.76 & 84.88 \color{blue}{(+1.04) }          & \textbf{59.64} \color{blue}{(+7.47) }  \\
+self-consistency  & \underline{69.10}  & \underline{57.71}          & \underline{63.94} \color{blue}{(+3.93) }         & 76.64          & 93.52 & 85.08 \color{blue}{(+1.16) }         & \underline{59.38} \color{blue}{(+7.21)}   \\ \toprule[2.0pt]      
\end{tabular}}
\vspace{-0.15in}
\caption{Accuracy of MLLMs on ChartBench, ChartQA, and ChartX. The best and second-best scores are highlighted in \textbf{bold} and \underline{underline}, respectively. The performance improvements over vanilla models are present in \color{blue}{brackets}.}
\vspace{-0.15in}
\label{eval_other_benchmark}
\end{table*}

\textbf{Fine-grained evaluation on questions with different lengths.}
We present the performance of MLLMs on questions with different lengths of function chains in Fig. \ref{fig:step_num}.
With the increase in lengths, MLLMs achieve lower accuracy since questions generally become difficult.
When adopting CoT, the performance for the questions with long function chains is improved, and the gap across lengths of function chains is minimized.
By deeply analyzing the accuracy of questions with different function chains, we surprisedly find that MLLMs can better handle some specific questions even with long function chains.
For example, we list the accuracy of three types of function chains in Fig. \ref{fig:acc_function_chain}: 1) object\_selection / filter / count; 2) object\_selection / value / object\_selection
 / min\_max / value / arithmetical\_operation; and 3) object\_selection / value / object\_selection
 / value / object\_selection / value / stat.
 The corresponding question examples are present in Table \ref{example_for_function_chain}.
 Even though the third one requires the longer reasoning chain to answer compared with the first one (7 vs. 2), InternVL-2.5-8B and Qwen-2-7B achieve better performance.

\textbf{Fine-grained evaluation on different question taxonomies.}
MLLMs achieve the significant performance difference in question taxonomies that possess different types of function chains.
We analyze the accuracy of some short function chains, including 1) object\_selection / value, 2) object\_selection / min\_max, and 3) object\_selection / filter, across MLLMs in Fig. \ref{fig:function combination}.
 The object\_selection / value stands for questions with value extraction, e.g., ``what is the value of object A?'' (See Example \ref{example 6}), object\_selection
 / min\_max stands for finding the minimum or maximum objects, e.g., ``what is the minimum value?'' (See Example \ref{example 5}), and object\_selection / filter stands for filtering unsatisfied objects, e.g., ``how many data points are larger than 925?'' (See Example \ref{example 1}).
We observe that InternVL-2.5-8B achieves a notably higher accuracy for the questions with the function chain object\_selection / value compared with GPT4o, even though it is the weaker model as presented in Table \ref{eval_chartcof}.
This indicates that InternVL-2.5-8B posseses better capabilities in value extraction on \textit{ChartCoF}.
However, GPT4o achieves significantly high accuracy on questions with object\_selection / filter, demonstrating that it masters the reasoning process of filtering data.
These fine-grained evaluations of reasoning capabilities on MLLMs provide effective guidance for model selection and training.

\subsection{Results on Existing Benchmarks}

Our proposed \textit{ChartCoF} can be used to enhance performance on existing benchmarks.
As shown in Table \ref{eval_other_benchmark}, after finetuning with \textit{ChartCoF}, InternVL-2.5-8B and Qwen2-VL-7B significantly improve the accuracy over ChartBench, ChartQA, and ChartX, with an improvement of 8.67\% for InternVL-2.5-8B in ChartBench and 7.47\% for Qwen2VL-7B in ChartX.
They outperform the existing MLLMs by a large margin and achieve state-of-the-art performance, demonstrating the effectiveness of \textit{ChartCoF} in enhancing the reasoning capabilities of existing MLLMs.
The self-consistency technique can further improve the performance of finetuned InternVL-2.5-8B and Qwen2VL-7B.

\subsection{Out-of-distribution Analysis}
To further demonstrate the effectiveness of \textit{ChartCoF}, we evaluate the OOD performance on unseen chart types and longer function chains.
We finetune InternVL-2.5-8B with only the regular charts (i.e., bar, line, and pie) and evaluate the accuracy performance in the extra test set of \textit{ChartCoF}, ChartBench, and ChartX (i.e., removing the regular charts from these benchmarks).
Results in Table \ref{tab:ood_chart_type} show that even with only the regular charts, InternVL-2.5-8B finetuned with \textit{ChartCoF} improves accuracy performance on the extra test set of all these three benchmarks, demonstrating that \textit{ChartCoF} can enhance the generalized reasoning capabilities on the unseen chart types.

\textit{ChartCoF} also enhances the generalization capabilities for longer function chains. 
We finetune InternVL-2.8-8B with the short-function-chain data (length of function chains $\leq 4,5,6$) and evaluate it on the long-function-chain test samples (length of function chains $\geq 7$).
Results on Table \ref{tab:ood_funstion_step} show that short-function-chain data significantly enhance the reasoning capabilities and improve accuracy on the long-function-chain test samples.
The OOD analysis demonstrates the effectiveness of \textit{ChartCoF} in boosting generalized reasoning capabilities, which attributes to the accurate and diverse CoT data generated by our proposed \textit{CoF} pipeline.

\begin{table}[]
    \centering
    \resizebox{\linewidth}{!}{
    \begin{tabular}{cccc}
    \midrule[1pt]
     Models    & ChartCoF & ChartBench & ChartX  \\ \midrule[1pt]
    InternVL-2.5-8B (direct answer)     &  42.84 & 41.73 & 42.64 \\
    InternVL-2.5-8B (CoT) & 40.00 & 29.07 & 38.19  \\
    InternVL-2.5-8B + ChartCoF (Reg.) & \textbf{51.04} & \textbf{46.40} & \textbf{43.25} \\
    \midrule[1pt]
    \end{tabular}}
    \vspace{-0.1in}
    \caption{Accuracy of MLLMs on benchmarks without regular chart types (bar, line, and pie).}
    \label{tab:ood_chart_type}
\end{table}

\begin{table}[]
\vspace{-0.1in}
    \centering
    \resizebox{\linewidth}{!}{
    \begin{tabular}{ccccc}
    \midrule[1pt]
 Length of function chains   & InternVL-2.5-8B (Vanilla) &  $\leq4$ & $\leq5$ & $\leq6$ \\ \midrule[1pt]
     ChartCoF ($\geq7$)    & 49.75 & 59.10 & 60.10 & \textbf{62.07}  \\ \midrule[1pt]
    \end{tabular}}
    \vspace{-0.1in}
    \caption{Accuracy of InternVL-2.5-8B with different training sets on the OOD test set of ChartCoF (i.e., length of function chains $\geq7$).}
    \label{tab:ood_funstion_step}
    \vspace{-0.15in}
\end{table}

\subsection{Model and Data Scalability}
\begin{figure}
    \centering
    \includegraphics[width=0.9\linewidth]{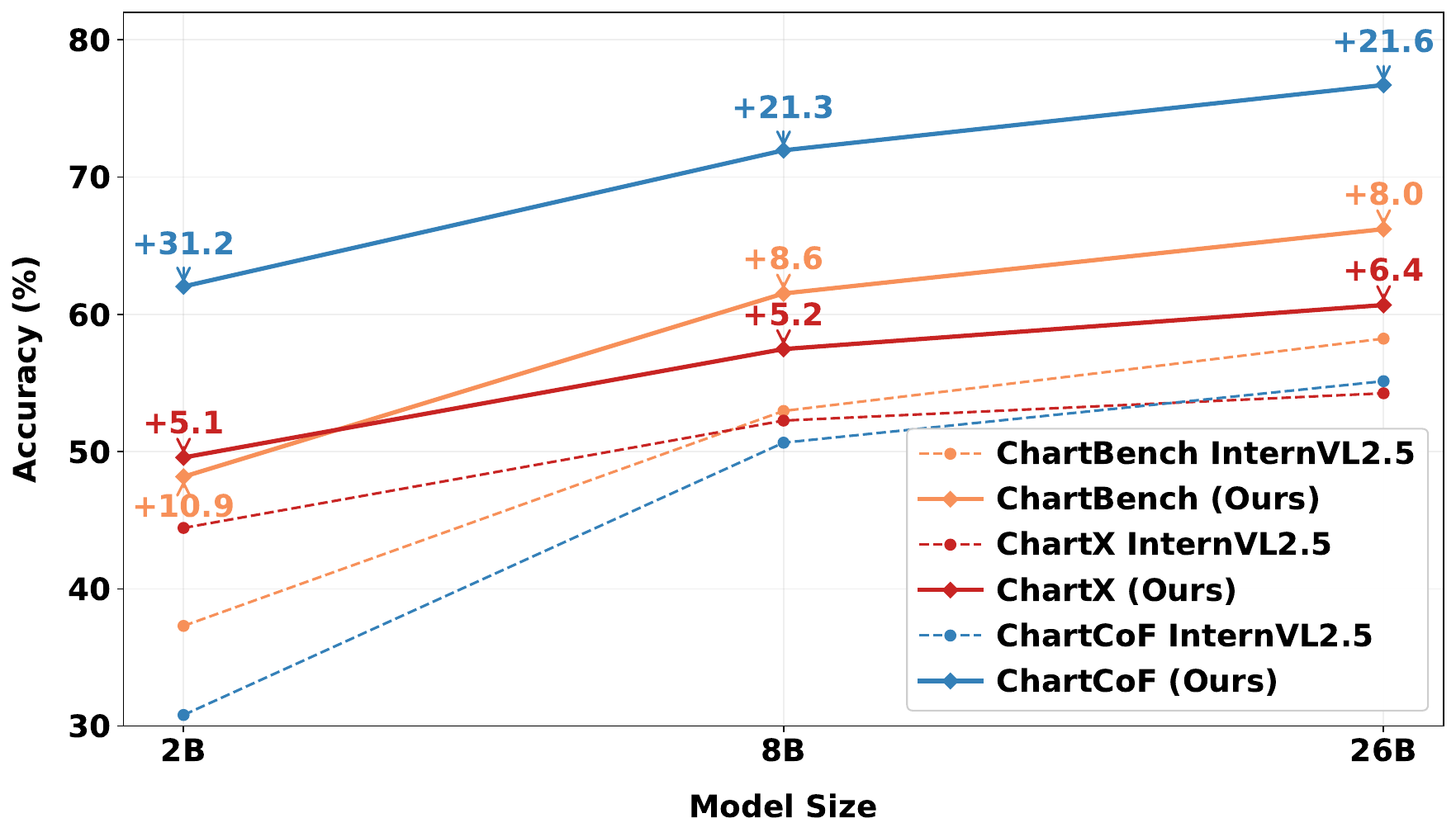}
    \vspace{-0.1in}
    \caption{Accuracy of InternVL2.5 series (2B, 8B, and 26B) on ChartBench, ChartX and ChartCoF.}
    \label{fig:model_scale}
\end{figure}

To further demonstrate the effectiveness of our proposed generation pipeline, we finetune different sizes of InternVL2.5 models and evaluate them on ChartBench, ChartX, and \textit{ChartCoF}.
The results in Fig. \ref{fig:model_scale} show that, 
with the increasing model parameters from 2B to 26B,
the accuracy of InternVL2.5 models keep increasing on ChartBench, ChartX, and \textit{ChartCoF}.
Meanwhile, after finetuning with the training set of our proposed \textit{ChartCoF}, all these three levels of InternVL2.5 models significantly outperform the base models without finetuning.
Remind that we utilize only Qwen2.5-32B-Instruct for data generation.
The notable improvement of the same-level model (i.e., InternVL2.5-26B) on benchmarks demonstrate that our data generation pipeline \textit{CoF} provide valuable supervision on data generation instead of only knowledge distillation from large models into small models.

\begin{figure}
    \centering
    \includegraphics[width=0.9\linewidth]{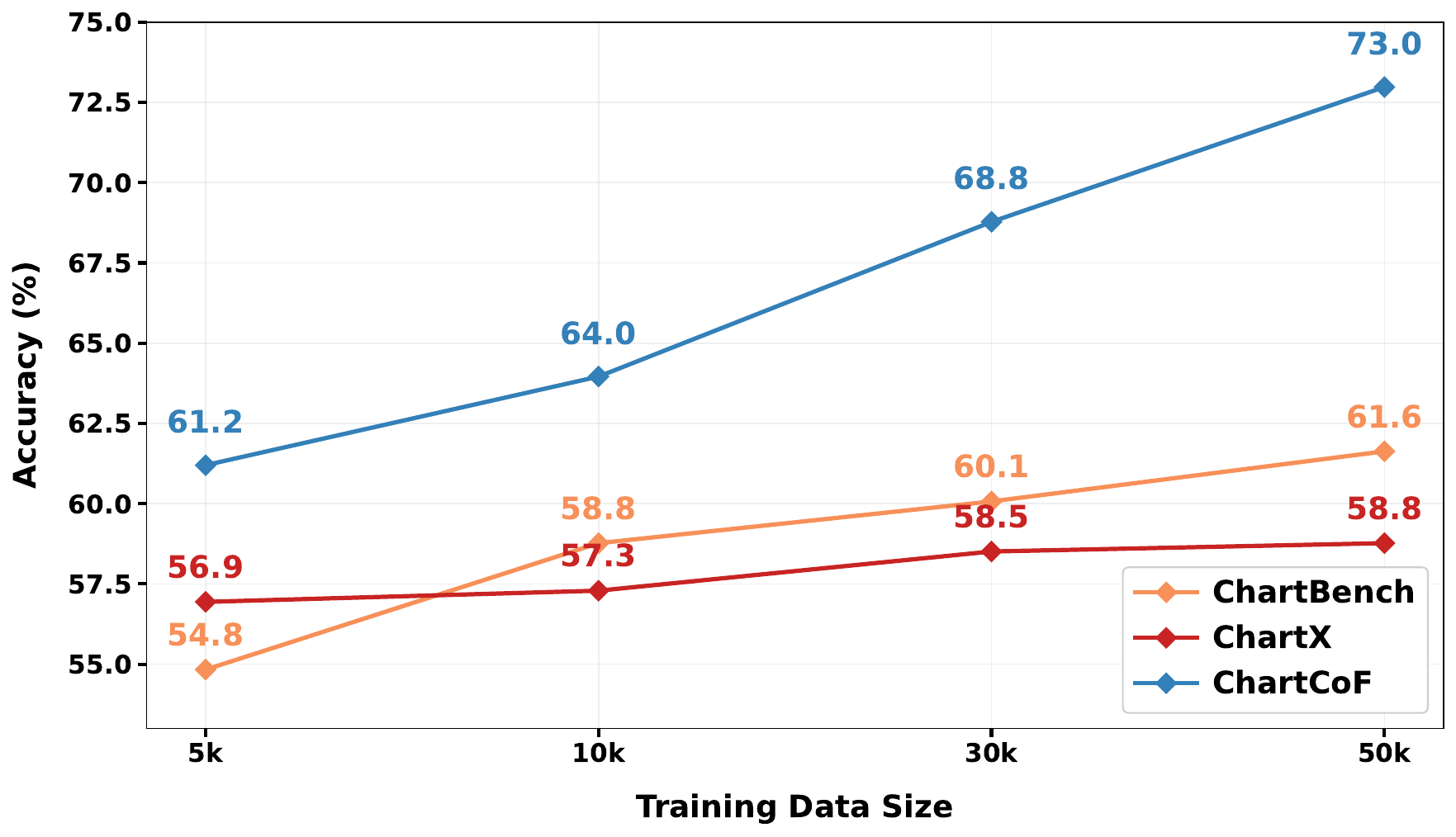}
    \vspace{-0.1in}
    \caption{Accuracy of InternVL2.5-8B on ChartBench, ChartX and ChartCoF.}
    \vspace{-0.1in}
    \label{fig:data_scale}
\end{figure}

To demonstrate the effectiveness of \textit{ChartCoF} on data scalability, we finetune InternVL2.5-8B with varying training data sizes on ChartBench, ChartX, and \textit{ChartCoF}.
The results in Fig. \ref{fig:data_scale} show that, with the increasing of the training data sizes, the accuracy of InternVL2.5-8B keeps increasing on these three benchmarks.
The effectiveness of \textit{ChartCoF} on data scalability reveals the potential of \textit{CoF} on generating larger scales of data to further improve the reasoning capabilities of MLLMs.

\section{Conclusion}
In this work, to overcome the scarcity of high-quality reasoning data for fine-grained evaluation and enhancement of chart reasoning capabilities, we proposed \textit{chain of functions (CoF)}, which utilized two key processes, including \textit{program-based functional discovery} and \textit{reverse linguistic CoT data synthesis}, to generate accurate and diverse reasoning data.
Employing \textit{CoF}, we introduced \textit{ChartCoF}, which enables the fine-grained evaluation on different reasoning questions and enhances the reasoning capabilities for chart understanding.
We believe that the ideas of \textit{functional discovery} and \textit{first exploration then task generation} in \textit{CoF} have the potential to extend to other step-wise tasks, such as mathematical Q\&A and graphical user interface tasks.

\section{Limitations}

We summarize the limitations of our work as below:
1) The current research emphasizes the critical role of chart data accuracy in the reasoning process for chart understanding. Consequently, we have chosen to represent charts using JSON data, rather than extracting charts directly from websites \cite{wangcharxiv, masry2022chartqa}. 
Despite our conscientious efforts to craft code templates specific to each chart type and the incorporation of diverse code libraries to increase the variety of charts, there remains a discernible difference between our synthesized charts and those that are naturally occurring on the internet. Future research could explore methodologies for the precise extraction of information from web-based charts or for the advancement of chart rendering techniques. Such innovations could narrow the existing chasm and enhance the reasoning proficiency of MLLMs.

2) Our approach leverages function chains as supervisory signals and employs LLMs as translators to generate reasoning data. Nevertheless, LLMs may still produce questions or rationales that are not entirely accurate on occasion. To ensure higher data quality, future efforts could focus on developing mechanisms to filter out these inaccuracies using state-of-the-art MLLMs. This would further refine the data generation process and enhance the reliability of the reasoning tasks performed by MLLMs.

\bibliography{custom}

\clearpage

\appendix

\begin{table}[h]
\resizebox{\linewidth}{!}{
\begin{tabular}{|cc|cc|cc|}
\hline
\multicolumn{2}{|c|}{\multirow{2}{*}{\textbf{Chart types}}}   & \multicolumn{2}{c|}{\textbf{Training set}} & \multicolumn{2}{c|}{\textbf{Test set}}    \\ \cline{3-6} 
\multicolumn{2}{|c|}{}                                        & \multicolumn{1}{c|}{\#charts}    & \#Q\&As & \multicolumn{1}{c|}{\#charts}   & \#Q\&As \\ \hline
\multicolumn{1}{|c|}{\multirow{6}{*}{Regular}} & Bar\_multi   & \multicolumn{1}{c|}{1819}         & 6050   & \multicolumn{1}{c|}{60}          & 165    \\ \cline{2-6} 
\multicolumn{1}{|c|}{}                         & Bar\_single  & \multicolumn{1}{c|}{1516}         & 6052   & \multicolumn{1}{c|}{60}          & 153    \\ \cline{2-6} 
\multicolumn{1}{|c|}{}                         & Bar\_stacked & \multicolumn{1}{c|}{1868}         & 6052   & \multicolumn{1}{c|}{57}          & 146    \\ \cline{2-6} 
\multicolumn{1}{|c|}{}                         & Line\_multi  & \multicolumn{1}{c|}{1541}         & 3050   & \multicolumn{1}{c|}{53}          & 123    \\ \cline{2-6} 
\multicolumn{1}{|c|}{}                         & Line\_single & \multicolumn{1}{c|}{1532}         & 3046   & \multicolumn{1}{c|}{62}          & 119    \\ \cline{2-6} 
\multicolumn{1}{|c|}{}                         & Pie          & \multicolumn{1}{c|}{655}          & 803    & \multicolumn{1}{c|}{37}          & 75     \\ \hline
\multicolumn{1}{|c|}{\multirow{13}{*}{Extra}}  & Radar        & \multicolumn{1}{c|}{104}          & 353    & \multicolumn{1}{c|}{25}          & 50     \\ \cline{2-6} 
\multicolumn{1}{|c|}{}                         & Rings        & \multicolumn{1}{c|}{514}          & 2050   & \multicolumn{1}{c|}{30}          & 50     \\ \cline{2-6} 
\multicolumn{1}{|c|}{}                         & Rose         & \multicolumn{1}{c|}{274}          & 1244   & \multicolumn{1}{c|}{25}          & 50     \\ \cline{2-6} 
\multicolumn{1}{|c|}{}                         & 3D-Bar       & \multicolumn{1}{c|}{611}          & 2054   & \multicolumn{1}{c|}{{23}} & 51     \\ \cline{2-6} 
\multicolumn{1}{|c|}{}                         & box          & \multicolumn{1}{c|}{627}          & 2050   & \multicolumn{1}{c|}{23}          & 50     \\ \cline{2-6} 
\multicolumn{1}{|c|}{}                         & funnel       & \multicolumn{1}{c|}{{964}} & 2054   & \multicolumn{1}{c|}{20}          & 50     \\ \cline{2-6} 
\multicolumn{1}{|c|}{}                         & heatmap      & \multicolumn{1}{c|}{696}          & 2055   & \multicolumn{1}{c|}{19}          & 50     \\ \cline{2-6} 
\multicolumn{1}{|c|}{}                         & area         & \multicolumn{1}{c|}{1007}         & 2050   & \multicolumn{1}{c|}{27}          & 50     \\ \cline{2-6} 
\multicolumn{1}{|c|}{}                         & bubble       & \multicolumn{1}{c|}{1120}         & 3107   & \multicolumn{1}{c|}{28}          & 80     \\ \cline{2-6} 
\multicolumn{1}{|c|}{}                         & node link    & \multicolumn{1}{c|}{1040}         & 2101   & \multicolumn{1}{c|}{34}          & 50     \\ \cline{2-6} 
\multicolumn{1}{|c|}{}                         & candlestick  & \multicolumn{1}{c|}{562}          & 2050   & \multicolumn{1}{c|}{17}          & 50     \\ \cline{2-6} 
\multicolumn{1}{|c|}{}                         & treemap      & \multicolumn{1}{c|}{989}          & 2052   & \multicolumn{1}{c|}{30}          & 50     \\ \cline{2-6} 
\multicolumn{1}{|c|}{}                         & multi-axes   & \multicolumn{1}{c|}{910}          & 2052   & \multicolumn{1}{c|}{18}          & 39     \\ \hline
\multicolumn{1}{|c|}{Total}                    &              & \multicolumn{1}{c|}{18349}        & 50329  & \multicolumn{1}{c|}{648}         & 1451   \\ \hline
\end{tabular}}
\caption{Detailed quantity statistics of each chart type for training and test sets of \textit{ChartCoF}.}
\label{quatity_chartcof}
\end{table}

\section{Data Splitting}
\label{data_splitting}
\textit{ChartCoF} dataset encompasses a total of 18,349 charts and 50,329 Q\&As in the training set, and 648 charts along with 1,451 Q\&As in the test set. 
The detailed quantity statistics for training and test sets of \textit{ChartCoF} are presented in Table \ref{quatity_chartcof}.
\textit{ChartCoF} is meticulously categorized into two distinct groups: Regular and Extra chart types. Within the Regular category, there are six different chart types, with Bar\_multi, Bar\_single, and Bar\_stacked leading in quantity for the training set, comprising 1,819, 1,516, and 1,868 charts, respectively. These three types also contribute to a substantial proportion of Q\&As, with each type exceeding 6,000 Q\&As. 
The Extra category encompasses a wider variety of 13 chart types for better generalization on chart types, which covers the chart types of existing benchmarks ChartBench \cite{xu2023chartbench} and ChartX \cite{xia2024chartx}.
Compared with Regular charts, the quantity of charts and Q\&As for each Extra type is slightly lower.
This comprehensive collection allows for robust training and effective evaluation of chart comprehension models, providing extensive coverage across a diverse range of chart types and complexity levels.

\begin{table}[h]
\resizebox{\linewidth}{!}{
    \centering
    \begin{tabular}{ccc}
    \toprule[1.5pt]
    & ChartQA  & ChartBench \\ \cmidrule(lr){1-1} \cmidrule(lr){2-3}
      w.o. rationale refinement   & 84.64 & 58.15  \\
      with rationale refinement   & 85.88   & 61.52 \\ \toprule[1.5pt]
    \end{tabular}}
    \caption{Ablation study of the effectiveness of rationale refinement for InternVL-2.5-8B on ChartQA and ChartBench.}
    \label{tab:ablation_rationale_refinement}
\end{table}

\begin{table*}[]
\vspace{-0.1in}
\resizebox{\textwidth}{!}{
\begin{tabular}{cccccccccc}
\toprule
\multirow{2}{*}{Dataset}    & \multicolumn{2}{c}{Chart properties}                                                                                                & \multicolumn{7}{c}{Q\&A properties}                                                                                                                                                                                                                                                 \\ \cmidrule(lr){2-3} \cmidrule(lr){4-10}
                      & \begin{tabular}[c]{@{}c@{}}\#Chart \\ Types\end{tabular} & \begin{tabular}[c]{@{}c@{}}Repre. \\ Format\end{tabular} & \begin{tabular}[c]{@{}c@{}}Func. \\ Usage\end{tabular} & \begin{tabular}[c]{@{}c@{}}Func. \\ Scal.\end{tabular} &  \begin{tabular}[c]{@{}c@{}}Rea. \\ Q\&A\end{tabular} & \begin{tabular}[c]{@{}c@{}}Lingui. \\ Rat.\end{tabular} & \begin{tabular}[c]{@{}c@{}}Func. Lengths \\ Eval.\end{tabular} & \begin{tabular}[c]{@{}c@{}}Ques. Tax. \\ Eval.\end{tabular} & Annotators           \\ \midrule 
    
ChartQA \cite{masry2022chartqa}               & 3             & Table                                                           & \redcheck                                                         & -                                                              &      \greencheck                                                                                            & \redcheck       & \redcheck    & \redcheck                                                            & Human                \\

SCI-CQA \cite{shen2024rethinking}             & 21             & -                                                               & \redcheck                                                         & -                                                              & \greencheck                                                                                                                & \redcheck       & \redcheck   & \redcheck                                                            & Human                \\
CharXiv \cite{wangcharxiv}             & -             & -                                                               & \redcheck                                                         & -                                                              & \greencheck                                                                                                                & \redcheck       & \redcheck   & \redcheck                                                            & Human                \\

MMC \cite{liu2024mmc}                  & 6             & Caption                                                         &  \redcheck                                                       & -                                                              & \greencheck                                                                                                                    & \greencheck   & \redcheck   & \redcheck                                                               & GPT-4                \\

ChartBench \cite{xu2023chartbench}           & 9             & Table                                                           & \redcheck                                                         & -                                                              & \redcheck                                                                                                              & \redcheck     & \redcheck   & \redcheck                                                              & GPT3.5               \\
ChartX \cite{xia2024chartx}               & 18            & Table                                                           & \redcheck                                                         & -                                                              & \greencheck                                                                                                                & \redcheck                                            & \redcheck   & \redcheck                       & GPT-4                \\
ChartLlama \cite{han2023chartllama}           & 10            & Table                                                           & \redcheck                                                         & -                                                              & \greencheck                                                                                                            & \greencheck    & \redcheck   & \redcheck                                                               & GPT-4                \\
ChartInstruct \cite{masry2024chartinstruct}        & -             & Table                                                           & \redcheck                                                         & -                                                              & \greencheck                                                                                           & \greencheck                                                      & \redcheck   & \redcheck             & GPT-4                \\
ChartGemma \cite{masry2024chartgemma}           & -             & -                                                               & \redcheck                                                         & -                                                              & \greencheck                                                                                    & \greencheck       & \redcheck   & \redcheck                                                            & Gemini Flash-1.5     \\
CHOPINLLM \cite{CHOPINLLM}            & 18            & JSON                                                            & \redcheck                                                         & -                                                              & \greencheck                                                                                                        & \greencheck                                        & \redcheck   & \redcheck                           & GPT-4                \\
REACHQA \cite{REACHQA}              & 10            & Code                                                            & \redcheck                                                         & -                                                              & \greencheck                                                                                                       & \greencheck                                         & \redcheck   & \redcheck                          & GPT4o                \\ \midrule \midrule
EvoChart \cite{huang2024evochart}             & 4             & Code                                                            & \greencheck                                                                                                                  & \redcheck                                                              & \greencheck                                                          & \redcheck                                                         & \redcheck   & \redcheck          & GPT-4                \\
PlotQA \cite{methani2020plotqa}              & 3             & Table                                                           & \greencheck                                                                                                               & \redcheck                                                              & \greencheck                                                          & \redcheck                                                                & -                    \\
ChartAst \cite{meng2024chartassisstant}             & 9             & Table                                                           & \greencheck                                                                                                                & \redcheck                                                              & \greencheck                                                          & \redcheck                         & \redcheck   & \redcheck                         & -                    \\
LAMENDA \cite{LAMENDA}              & 3             & Table                                                           & \greencheck                                                                                                               & \redcheck                                                              & \greencheck                                                          & \redcheck                          & \redcheck   & \redcheck                                         & -                    \\
\textbf{ChartCoF (ours)  }                & 19            & JSON                                                            & \greencheck                                                                                                                   & \greencheck                                                              & \greencheck                                                          & \greencheck & \greencheck & \greencheck                                                                & Qwen2.5-32B \\ \toprule
\end{tabular}}
\vspace{-0.1in}
\caption{Comparison between \textit{ChartCoF} and existing chart-related datasets. Abbreviations: Repre.=Representation, Scal.=Scalability, Rea.=Reasoning, Lingui.=Linguistic, Rat.=Rationale, Func.=Function Ques.=Questions Tax.=taxonomy. \textit{ChartCoF} enables accurate and diverse reasoning data via scalable function usage and additionally provides fine-grained evaluation on different function lengths and question taxonomies.}
\vspace{-0.1in}
\label{dataset_comparison}
\end{table*}

\section{Dataset Comparison}
\label{appendix_dataset_comparison}
We provide a detailed comparison between \textit{ChartCoF} and existing datasets from the aspects of evaluation and quality of training data, as presented in Table \ref{dataset_comparison}.
ChartQA \cite{masry2022chartqa}, SCI-CQA \cite{shen2024rethinking}, and ChrXiv \cite{wangcharxiv} provide reasoning questions with the charts from webs, where the questions are annotated by humans.
Despite the delicate charts and reasoning questions, the barier of human annotations makes them hard to scale to the training set.
Besides, these benchmarks only provide a coarse evaluation with an accuracy metric.
To scalably generate instruction data, some studies, including MMC \cite{liu2024mmc}, ChartBench \cite{xu2023chartbench}, ChartX \cite{xia2024chartx}, ChartLlama \cite{han2023chartllama}, ChartInstruct \cite{masry2024chartinstruct}, ChartGemma \cite{masry2024chartgemma}, CHOPINLLM \cite{CHOPINLLM}, and REACHQA \cite{REACHQA}, have utilized extremely large (M)LLMs to generate reasoning instructions.
However, the autoregressive generation and fix-pattern prompts for generation limit precision and diversity of generated instructions.
Although EvoChart \cite{huang2024evochart}, PlotQA \cite{methani2020plotqa}, ChartAst \cite{meng2024chartassisstant}, and LAMENDA \cite{LAMENDA} have manually set up program or function templates to ensure the precision of instructions, the predefined templates still suffer from the low diversity of instructions, and they cannot provide the linguistic rationales for enhancing the reasoning capabilities.
Overall, compared with existing datasets, \textit{ChartCoF} provides more diverse and accurate reasoning data for enhancing the reasoning capabilities and fine-grained evaluation on the varying question taxonomies.

\section{Effectiveness of Rationale Refinement}
\label{ablation_rationale_refinement}
To enhance the effectiveness of rationale refinement, we conduct the ablation study for it.
We finetune InternVL-2.5-8B using the same number of samples without rationale refinement and evaluate it on ChartQA and ChartBench.
Results on Table \ref{tab:ablation_rationale_refinement} show that after rationale refinement, the performance of InternVL-2.5-8B is improved on ChartQA and ChartBench.
A comparison example is presented in Example \ref{example_rationale_refinement}.
Before rationale refinement, the rationale manuscript is still function-like and redundant due to the translation from function chains.
After refinement, the rationale becomes more fluent and information-intensive, which better aligns with the existing MLLMs and enhances their reasoning capabilities.


\subsection{Example of rationale refinement}
\label{example_rationale_refinement}
\begin{example}[Example of rationale refinement]
\tiny{
\begin{lstlisting}[breaklines, basicstyle=\ttfamily]
Question: 
According to the chart, which legend is associated with the light green data points representing satisfaction scores from customer feedback across different service providers?

Rationale manuscript:
First, we select all the data points that are represented in light green on the chart, which corresponds to 'Client Feedback'. These points are from 'Provider A' to 'Provider I'. Then, we identify the legend associated with these data points, which is 'Client Feedback'. Final answer: Client Feedback

After rationale refinement:
The data points shown in light green on the chart represent 'Client Feedback' from different service providers, specifically from Provider A to Provider I. The legend associated with these light green data points is 'Client Feedback'. Final answer: Client Feedback

\end{lstlisting}}
\end{example}

\section{JSON Templates}
\label{json_template}
Without specific statements on chart types, the general JSON templates for charts are presented in Appendix \ref{json_elements}.
The JSON templates for box, candlestick, and node link charts are presented in Appendices \ref{json_elements_box}, \ref{json_elements_candlestick}, and \ref{json_elements_node_link}, respectively.

\subsection{JSON Elements for Charts}
\label{json_elements}
\begin{example}[JSON elements]
\tiny{
\begin{lstlisting}[breaklines, basicstyle=\ttfamily]
{
"title": {title},
"x_label": {x_label},
"y_label": {y_label},
"type": {type},
"legend_num": {legend_num},
"legends": [legend list],
"group_num": {group_num},
"groups": [group list],
"colors": {colors},
"data_points":
    {
    "group1": {
        legend1: xxxx, 
        legend2: xxxx
        },
    },
"legend_colors":
    {
    "legend1": "color1",
    "legend2": "color2",
    },
}
\end{lstlisting}}
\end{example}

\clearpage

\subsection{JSON Elements for Box}
\label{json_elements_box}
\begin{example}[JSON elements for box]
\tiny{
\begin{lstlisting}[breaklines, basicstyle=\ttfamily]
{
"title": {title},
"x_label": {x_label},
"y_label": {y_label},
"type": {type},
"legend_num": {legend_num},
"legends": [legend list],
"group_num": {group_num},
"groups": [group list],
"colors": {colors},
"legend_colors":
    {
    "legend1": "color1",
    "legend2": "color2"
    },
"median": {"legend1": xxx},
"first_quartile": {"legend1": xxx},
"third_quartile": {"legend1": xxx},
"minimum_values": {"legend1": xxx},
"maximum_values": {"legend1": xxx},
"outlier_values": {"legend1": xxx}
}
\end{lstlisting}}
\end{example}

\subsection{JSON Elements for Candlestick}
\label{json_elements_candlestick}
\begin{example}[JSON elements for Candlestick]
\tiny{
\begin{lstlisting}[breaklines, basicstyle=\ttfamily]
{
"title": {title},
"x_label": {x_label},
"y_label": {y_label},
"type": {type},
"legend_num": {legend_num},
"legends": [legend list],
"group_num": {group_num},
"groups": [group list],
"colors": {colors},
"legend_colors":
    {
    "legend1": "color1",
    "legend2": "color2"
    },
"opening_price": {"legend1": xxx},
"closing_price": {"legend1": xxx},
"highest_price": {"legend1": xxx},
"lowest_price": {"legend1": xxx}
}
\end{lstlisting}}
\end{example}

\subsection{JSON Elements for Node Link}
\label{json_elements_node_link}
\begin{example}[JSON elements for Node Link]
\tiny{
\begin{lstlisting}[breaklines, basicstyle=\ttfamily]
{
"title": {title},
"x_label": {x_label},
"y_label": {y_label},
"type": {type},
"legend_num": {legend_num},
"legends": [legend list],
"group_num": {group_num},
"groups": [group list],
"colors": {colors},
"data_points":
    {
    "group1": {legend1: [pointed_object_list_1]},
    "group2": {legend1: [pointed_object_list_1]},
    },
"legend_colors":
    {
    "legend1": "color1",
    "legend2": "color2"
    }
}
\end{lstlisting}}
\end{example}

\section{Prompts Usage}
\label{prompt_json_generation_evolement}

The prompts for JSON seed generation and JSON evolement are presented in Prompts \ref{prompt_json_generation} and \ref{prompt_json_evolement}, respectively.
The prompts for rationale generation, question generation, and rationale refinement are presented in Prompts \ref{prompt_rationale_generation}, \ref{prompt_question_generation}, and \ref{prompt_rationale_refinement}, respectively.
The prompt for answer extraction is presented in Prompt \ref{prompt_answer_extraction}.

\onecolumn

\subsection{Prompt for JSON Seed Generation}

\label{prompt_json_generation}
\begin{example}[Prompt for JSON Seed Generation]
\tiny{
\begin{lstlisting}[breaklines, basicstyle=\ttfamily]
You are a language model tasked with generating augmented datasets to train machine learning models for chart understanding. These models need to be exposed to various chart configurations, data patterns, and types to perform accurately in diverse scenarios.
Given a JSON template that contains the basic information for a chart, your task is to fill in the missing details to generate a new JSON data.

Instructions:
1. The title, type, colors, legend_num, and group_num are given, and you need to add x_label, y_label, data_points, legends, and groups.
2. Ensure that the augmented data is diverse and realistic.
3. Maintain the structure and integrity of the original data.
4. According to the legend_num and group_num, generate the corresponding legends and groups.
5. Assign the colors in "colors" to each legend.

The original JSON data is as follows:
{JSON element file}

The output format should be: JSON Data 1: <Augmented JSON data 1>.
Only output the augmented JSON data that can be directly used to generate the chart.
\end{lstlisting}}
\end{example}

\subsection{Prompt for JSON Evolement}

\label{prompt_json_evolement}
\begin{example}[Prompt for JSON Evolement]
\tiny{
\begin{lstlisting}[breaklines, basicstyle=\ttfamily]
You are a language model tasked with generating augmented datasets to train machine learning models for chart understanding. These models need to be exposed to various chart configurations, data patterns, and types to perform accurately in diverse scenarios.
Given a JSON script, your task is to correct and enrich the JSON data to generate a new JSON data.

Instructions:
1. Title: change the title of the chart to make it more descriptive and informative to the type.
2. x_label and y_label: change the x_label and y_label to make them more compatible to the title.
3. Data points: if the data points are not satisfying with the type, title, x_label, and y_label, recorrect the data points to make them more realistic. Your can add some noise to the data points to make them more diverse. 
4. Legends: keep the legend_num unchanged. Change the legends to make them more informative and diverse.
5. Groups: change the group_num and groups to make them more diverse and informative. Make sure that the length of groups is the same as the group_num.
6. Colors: change the colors of the chart to make it more visually appealing and informative. Make sure that the colors are different and sampled from {color_list}, and the color number should be the same as the legend_num.
7. Save the new JSON data as {data_save_path}.

The original JSON data is as follows:
{json_data}

The output format should be: JSON Data 1: <Augmented JSON data 1>.
Only output the augmented JSON data that can be directly used to generate the chart.
\end{lstlisting}}
\end{example}


\clearpage

\subsection{Prompt for Rationale Generation}

\label{prompt_rationale_generation}
\begin{example}[Prompt for Rationale Generation]
\tiny{
\begin{lstlisting}[breaklines, basicstyle=\ttfamily]
You are an AI assistant specialized in translating technical reasoning processes into clear, natural language explanations for chart reasoning. You will be given the JSON data of the chart and a structured description of a chart understanding process, which includes inputs, functions, and outputs. Your task is to convert this structured information into a coherent, easy-to-understand paragraph.

Please follow these guidelines to generate rationale with natural language:

1. Before the reasoning process, different legends, categories, or colors are sampled. You should take them as conditions. 
2. The reasoning processes should be related to chart understanding.
3. Describe the purpose and action of each function in simple terms.
4. When the function is related to the values of data, list all the values of the data.
5. When the function is related to the numerical calculation, you should provide calculation process and the final answer uising numerical operations, e.g., A + B = D, A - B = D, A * B = D, A / B = D, (A + B + C) / 3 = D etc.
6. Some functions that related to position, like left, right, top, bottom are used to render the data using the position information. You should emphasize the position in the rationale.
7. Some functions that related to colors are used to render the data using the color information. You should emphasize the color in the rationale. 
8. If the function is specific to some charts, like bar, line, and pie, you should mention the chart type.
9. The final output should be the final answer.
{addition_prompt}

The JSON data of the chart:

{json_str}

Here's the structured process description: {structured process description}

Only transfer the structured process to a natural languages in short sentances.
The output format should be like:
Reasoning process: [Your reasoning process], Final answer: [Your final answer]
\end{lstlisting}}
\end{example}

\subsection{Prompt for Question Generation}

\label{prompt_question_generation}
\begin{example}[Prompt for Question Generation]
\tiny{
\begin{lstlisting}[breaklines, basicstyle=\ttfamily]
You are an AI assistant specialized in generating questions for chart reasoning. You will be given the JSON data of the chart, the reasoning process, and its corresponding structured description of a chart understanding process, which includes inputs, functions, and outputs. Your task is to generate a coherent, easy-to-understand question that can be answered by the reasoning process.

Please follow these guidelines:

1. Your question should follow the structured process of the chart.
2. The question can be answered by the structured process.
3. During the reasoning process, different legends, categories, or colors are used to refer data. You should consider them as conditions and emphasize them in the question. 
4. If the rationale contains the color, you should take it as a condition and emphasize it in the question.
5. If the rationale contains the position information, like upper, bottom, left, and right, you should take them as conditions and emphasize them in the question.
6. The question should consider all the functions in the structured process.
7. For the length of structured process description is longer than 4 steps, you can first illustrate the conditions to get the data and then ask the question. You can use the patterns like "If we get a value through xxx and get another value through yyy, what/how/...?".
8. For the length of structured process description is shorter than 4 steps, you can directly ask the question.
9. Do not appear the important intermediate values or information (categories, legends, and colors) of data in the question directly since they need to be calculated by the question.

The JSON data of the chart:

{json_data}

Here's the structured process description:
{structured process description}

Here's the reasoning process in short sentences:
{rationale}

Please generate a question that can be answered by the structured process and reasoning process.
The output format should be Question: [Your question]
\end{lstlisting}}
\end{example}

\subsection{Prompt for Rationale Refinement}

\label{prompt_rationale_refinement}
\begin{example}[Prompt for Rationale Refinement]
\tiny{
\begin{lstlisting}[breaklines, basicstyle=\ttfamily]
You are an AI assistant specialized in answering questions. You are given a structured process description, the rationale manuscript, and the question. You need to answer the question according to the structured process description and the rationale manuscript.

The structured process description is as follows:
{structured process description}

The question is as follows:
{question}

The rationale manuscript is as follows:
{rationale}

You should answer the question under the following constraints:
1. Imagine that you are answering the question about charts in a real-world scenario. You answer should be related to the chart understanding.
2. You should first answer the question step by step to generate rationale by taking the structured process description as evidence, but "structured process description" should not be mentioned in the answer.
3. The answer should be consistent with the structured process description. 
4. You should keep the rationale fluent, understandable, and concise.
5. You can fuse the structured process description and the rationale manuscript to make the answer more understandable and concise.
6. You should remove the personal pronoun and focus on the elements that are related to the question.
7. If there are some numerical values in the reasoning processes, try to maintain the numerical values in the answer to make the answer more accurate.
8. If there are calculations in the reasoning processes, you should use the mathematical symbols in the natural language description to improve the readability.

The output format should be like:
Rewritten rationale: [Your rewritten rationale], Final answer: [Your final answer]
\end{lstlisting}}
\end{example}

\subsection{Prompt for Answer Extraction}

\label{prompt_answer_extraction}
\begin{example}[Prompt for Answer Extraction]
\tiny{
\begin{lstlisting}[breaklines, basicstyle=\ttfamily]
Please extract the answer from the model response and type it.

Note:
1. The responses may be a phrase, a number, or a sentence.
2. If the content of the responses is not understandable, return "FAILED".
3. If the content of the responses is understandable, extract the numerical value from it.
4. If the responses is a yes or no judgment, return yes or no.
5. If the answer contains a unit, please exclude the unit and only return the numerical value.

Special requirements: ** Only numbers, short texts, "FAILED", or yes/no are allowed to be returned for each response, please do not return anything else! **

Please read the following example. 

Question 1: Which number is missing?
Model response: The number missing in the sequence is 14.

Question 2: What is the fraction of females facing the camera?
Model response: The fraction of females facing the camera is 0.6, which means that six out of ten females in the group are facing the camera.

Question 3: How much money does Luca need to buy a sour apple candy and a butterscotch candy? (Unit: $)
Model response: Ax00 Ax00 Ax00 Ax00 Ax00 Ax00 Ax00 Ax00 Ax00 Ax00 Ax00. 

Question 4: In the chart titled \"Quarterly Sales Breakdown by Product Category\", if we identify the product category with the second lowest sales value for Q1 2023, what is the color associated with that category?
Model response: The product category with the second lowest sales value for Q1 2023 is Jewelry. The color associated with that category is gray.

Question 5: Which month shows the smallest difference in visitors between mobile devices and desktop devices?
Model response: The difference in visitors between mobile devices and desktop devices is the smallest in Apr.

Your answer: 
14
0.6
FAILED
gray
Apr

Question: {}
Model response: {}
Expected answer:
\end{lstlisting}}
\end{example}

\clearpage

\twocolumn

\section{Object Functions and Value Functions}
\label{appendix_function}
We adopt 6 selection methods for object selection and set up 99 object functions and 8 value functions in experiments.
The detailed functions for object selection are presented in Table \ref{object_selection}.
The object functions for box, candlestick, and node link charts are presented in Tables \ref{object_function_box}, \ref{object_function_candlestick}, \ref{object_function_node_link}, respectively.
Without specific statements on chart types, the general object functions for charts are presented in Table \ref{object_function_details}.
The value functions are presented in Table \ref{value_function}.

We categorize the functions into several function taxonomies according to their purpose for statistical analysis.
The statistics of the function taxonomy are presented in Table \ref{stat_function_tax}.
Among them, the most frequent function taxonomy in the test set of \textit{ChartCoF} is ``value'', which stands for the value extraction functions.
This is because value extraction is very common in the reasoning process of chart understanding, and numerous function chains also contain the value extraction function.

\begin{table*}[h]
\centering
\resizebox{0.6\linewidth}{!}{
\begin{tabular}{|l|l|}
\hline
Object selection           & Description                                             \\ \hline
one\_object\_selection     & Select one object using a  group name and a legend name \\ \hline
group\_selection           & Select partial objects using a group name               \\ \hline
legend\_selection          & Select partial objects using a legend name              \\ \hline
color\_selection           & Select partial objects using a color                    \\ \hline
color\_group\_selection & Select one object using a  group name and a color       \\ \hline
all\_object\_selection     & Select all the objects of the chart                     \\ \hline
\end{tabular}}
\caption{Overview of object selection.}
\vspace{-0.2in}
\label{object_selection}
\end{table*}


\begin{table*}[h]
\resizebox{\linewidth}{!}{
\begin{tabular}{|l|l|l|}
\hline
Function taxonomy       & Description                                                                      & Percentage \\ \hline
value                   & The functions related to value extraction                                        & 43.36\%    \\ \hline
text\_information       & The functions related to text information of charts                              & 4.76\%     \\ \hline
count                   & The functions related to counting                                                & 3.54\%     \\ \hline
min\_max                & The functions related to maximum or minimum values                               & 17.61\%    \\ \hline
arithmetical\_operation & The functions related to arithmetical operation                                  & 6.88\%     \\ \hline
compare                 & The functions related to comparison                                              & 3.63\%     \\ \hline
stat                    & The functions related to statistics                                              & 8.18\%     \\ \hline
filter                  & The functions related to filtering unsatisfied objects                           & 4.93\%     \\ \hline
if\_match\_condition    & The functions related to assessing if the objects or values match the conditions & 2.09\%     \\ \hline
exclude\_objects        & The functions related to excluding the objects with some conditions              & 0.36\%     \\ \hline
position                & The functions related to the position of objects                                 & 4.57       \\ \hline
\end{tabular}}
\caption{The percentage for each function taxonomy in the test set of \textit{ChartCoF}.}
\label{stat_function_tax}
\end{table*}

\begin{table*}[h]
\resizebox{\linewidth}{!}{
\begin{tabular}{|l|l|l|l|}
\hline
\textbf{Function taxtonomy}            & Functions                               & description                                                             & Input conditions                                                   \\ \hline
\multirow{3}{*}{text\_information} & color\_of\_objects                      & Return the color of the object.                                         & len(obejcts)=1                                                     \\ \cline{2-4} 
                                   & groups\_of\_object                   & Return the groups of the object.                                        & one\_object\_selection not in function chain                       \\ \cline{2-4} 
                                   & legends\_of\_object                     & Return the legend of the object.                                        & one\_object\_selection and legend\_selection not in function chain \\ \hline
\multirow{7}{*}{value}             & median\_of\_objects                     & Return the median value of the boxplot.                                 & -                                                                  \\ \cline{2-4} 
                                   & first\_quartile\_of\_objects            & Return the first quartile value of the boxplot.                         & -                                                                  \\ \cline{2-4} 
                                   & third\_quartile\_of\_objects            & Return the third quartile value of the boxplot.                         & -                                                                  \\ \cline{2-4} 
                                   & maximum\_value\_without\_outliers       & Return the maximum value of the boxplot without outliers.               & -                                                                  \\ \cline{2-4} 
                                   & minimum\_value\_without\_outliers       & Return the minimum value of the boxplot without outliers.               & -                                                                  \\ \cline{2-4} 
                                   & interquartile\_range\_of\_box           & Return the interquartile range of the boxplot.                          & len(obejcts)=1                                                     \\ \cline{2-4} 
                                   & outlier\_values\_of\_objects            & Return the outlier values of the boxplot.                               & len(obejcts)=1                                                     \\ \hline
\multirow{10}{*}{min\_max}         & max\_median\_object                     & Return the object with the maximum median value of the boxplot.         & len(obejcts)\textgreater{}1                                        \\ \cline{2-4} 
                                   & min\_median\_object                     & Return the object with the minimum median value of the boxplot.         & len(obejcts)\textgreater{}1                                        \\ \cline{2-4} 
                                   & max\_maximum\_object\_without\_outliers & Return the object with the maximum maximum value of the boxplot.        & len(obejcts)\textgreater{}1                                        \\ \cline{2-4} 
                                   & min\_maximum\_object\_without\_outliers & Return the object with the minimum maximum value of the boxplot.        & len(obejcts)\textgreater{}1                                        \\ \cline{2-4} 
                                   & max\_minimum\_object\_without\_outliers & Return the object with the maximum minimum value of the boxplot.        & len(obejcts)\textgreater{}1                                        \\ \cline{2-4} 
                                   & min\_minimum\_object\_without\_outliers & Return the object with the minimum minimum value of the boxplot.        & len(obejcts)\textgreater{}1                                        \\ \cline{2-4} 
                                   & max\_first\_quartile\_object            & Return the object with the maximum first quartile value of the boxplot. & len(obejcts)\textgreater{}1                                        \\ \cline{2-4} 
                                   & min\_first\_quartile\_object            & Return the object with the minimum first quartile value of the boxplot. & len(obejcts)\textgreater{}1                                        \\ \cline{2-4} 
                                   & max\_third\_quartile\_object            & Return the object with the maximum third quartile value of the boxplot. & len(obejcts)\textgreater{}1                                        \\ \cline{2-4} 
                                   & min\_third\_quartile\_object            & Return the object with the minimum third quartile value of the boxplot. & len(obejcts)\textgreater{}1                                        \\ \hline
count                              & num\_of\_outliers                       & Return the number of outliers of the boxplot.                           & len(obejcts)=1                                                     \\ \hline
\multirow{4}{*}{position}          & leftmost\_box                           & Return the leftmost box in the boxplot.                                 & len(obejcts)\textgreater{}1                                        \\ \cline{2-4} 
                                   & rightmost\_box                          & eturn the rightmost box in the boxplot.                                 & len(obejcts)\textgreater{}1                                        \\ \cline{2-4} 
                                   & upper\_box                              & Return the upper box in the boxplot.                                    & len(obejcts)\textgreater{}1                                        \\ \cline{2-4} 
                                   & bottom\_box                             & Return the bottom box in the boxplot.                                   & len(obejcts)\textgreater{}1                                        \\ \hline
\end{tabular}}
\vspace{-0.1in}
\caption{Overview of object functions for box charts.}
\vspace{-0.2in}
\label{object_function_box}
\end{table*}

\begin{table*}[h]
\centering
\tiny{
\resizebox{0.8\linewidth}{!}{
\begin{tabular}{|l|l|l|l|}
\hline
\textbf{Function taxtonomy}   & Functions                 & description                                     & Input conditions            \\ \hline
text\_information         & legends\_of\_object       & Return the legend of the object.                & len(obejcts)=1              \\ \hline
\multirow{4}{*}{value}    & high\_price\_of\_object   & Return the high price of the object.            & len(obejcts)=1              \\ \cline{2-4} 
                          & low\_price\_of\_object    & Return the low price of the object.             & len(obejcts)=1              \\ \cline{2-4} 
                          & open\_price\_of\_object   & Return the open price of the object.            & len(obejcts)=1              \\ \cline{2-4} 
                          & close\_price\_of\_object  & Return the close price of the object.           & len(obejcts)=1              \\ \hline
\multirow{8}{*}{min\_max} & max\_high\_price\_object  & Return the object with the maximum high price.  & len(obejcts)\textgreater{}1 \\ \cline{2-4} 
                          & min\_high\_price\_object  & Return the object with the minimum high price.  & len(obejcts)\textgreater{}1 \\ \cline{2-4} 
                          & max\_low\_price\_object   & Return the object with the maximum low price.   & len(obejcts)\textgreater{}1 \\ \cline{2-4} 
                          & min\_low\_price\_object   & Return the object with the minimum low price.   & len(obejcts)\textgreater{}1 \\ \cline{2-4} 
                          & max\_open\_price\_object  & Return the object with the maximum open price.  & len(obejcts)\textgreater{}1 \\ \cline{2-4} 
                          & min\_open\_price\_object  & Return the object with the minimum open price.  & len(obejcts)\textgreater{}1 \\ \cline{2-4} 
                          & max\_close\_price\_object & Return the object with the maximum close price. & len(obejcts)\textgreater{}1 \\ \cline{2-4} 
                          & min\_close\_price\_object & Return the object with the minimum close price. & len(obejcts)\textgreater{}1 \\ \hline
\end{tabular}}}
\vspace{-0.1in}
\caption{Overview of object functions for candlestick charts.}
\label{object_function_candlestick}
\vspace{-0.2in}
\end{table*}

\begin{table*}[h]
\tiny{
\resizebox{\linewidth}{!}{
\begin{tabular}{|l|l|l|l|}
\hline
\textbf{Function taxonomy}               & Functions                  & description                                                                & Input conditions \\ \hline
text\_information                     & legend\_of\_objects        & Return the legends (name) of the objects                                   & -                \\ \hline
\multirow{3}{*}{filter}               & targets\_of\_object        & Return the target objects that the object points to with an arrow          & len(obejcts)=1   \\ \cline{2-4} 
                                      & sources\_of\_object        & Return the sourced objects that are pointed by the object with an arrow    & len(obejcts)=1   \\ \cline{2-4} 
                                      & connected\_objects         & Return the conntected objects that are connected to the object with a line & len(obejcts)=1   \\ \hline
\multirow{3}{*}{if\_match\_condition} & if\_object\_point\_to\_A   & Return whether the object point to \{A\} with an arrow                     & len(obejcts)=1   \\ \cline{2-4} 
                                      & if\_object\_pointed\_by\_A & Return whether the object is pointed by \{A\} with an arrow                & len(obejcts)=1   \\ \cline{2-4} 
                                      & if\_object\_connect\_to\_A & Return whether the object is connected to \{A\}                            & len(obejcts)=1   \\ \hline
\end{tabular}}}
\vspace{-0.1in}
\caption{Overview of objective functions for node link charts.}
\label{object_function_node_link}
\vspace{-0.2in}
\end{table*}

\begin{table*}[h]
\resizebox{\linewidth}{!}{
\begin{tabular}{|l|l|l|l|}
\hline
\textbf{Function taxtonomy}                  & Functions                      & description                                                                                               & Input conditions           \\ \hline
\multirow{2}{*}{stat}                   & sum\_of\_values                & Return the sum of the values of data: A + B + C.                                                          & len(values)\textgreater{}1 \\ \cline{2-4} 
                                         & mean\_of\_values               & Return the mean of the values of data: (A + B + C) / len = D / len.                                       & len(values)\textgreater{}1 \\ \hline
\multirow{6}{*}{arithmetical\_operation} & median\_of\_values             & Return the median value of data.                                                                          & len(values)\textgreater{}1 \\ \cline{2-4} 
                                         & A\_minus\_B                    & Return A - B.                                                                                             & len(values)=2              \\ \cline{2-4} 
                                         & difference\_between\_A\_and\_B & Return the difference between two data: | A - B |.                                                        & len(values)=2              \\ \cline{2-4} 
                                         & A\_multiply\_B                 & Return the product of two data: A * B.                                                                    & len(values)=2              \\ \cline{2-4} 
                                         & A\_divided\_by\_B              & Return the division of two data: A / B.                                                                   & len(values)=2              \\ \cline{2-4} 
                                         & multiply\_constant             & Return the value multiplied by a constant \{constant\}: A * constant.                                     & len(values)=1              \\ \hline
\multirow{2}{*}{compare}                 & A\_is\_larger\_than\_B         & Return True if the value of the first data is larger than the value of the second data: A \textgreater B. & len(values)=2              \\ \cline{2-4} 
                                         & A\_is\_smaller\_than\_B        & Return True if the value of the first data is smaller than the value of the second data: A \textless B.   & len(values)=2              \\ \hline
\end{tabular}}
\vspace{-0.1in}
\caption{Overview of value functions.}
\label{value_function}
\vspace{-0.2in}
\end{table*}

\begin{table*}[h]
\resizebox{\linewidth}{!}{
\begin{tabular}{|l|l|l|l|}
\hline
\textbf{Function taxtonomy}               & Functions                                          & description                                                                                                                                                                                   & Input conditions                                                                                                                                                                        \\ \hline
\multirow{8}{*}{min\_max}             & max\_one\_object                                   & Return the data with the maximum value \{value\}.                                                                                                                                             & len(obejcts)\textgreater{}1                                                                                                                                                             \\ \cline{2-4} 
                                      & min\_one\_object                                   & Return the data with the minimum value \{value\}.                                                                                                                                             & len(obejcts)\textgreater{}1                                                                                                                                                             \\ \cline{2-4} 
                                      & max\_two\_objects                                  & Return the two data with the maximum values \{value\}.                                                                                                                                        & len(obejcts)\textgreater{}2                                                                                                                                                             \\ \cline{2-4} 
                                      & min\_two\_objects                                  & Return the two data with the minimum values \{value\}.                                                                                                                                        & len(obejcts)\textgreater{}2                                                                                                                                                             \\ \cline{2-4} 
                                      & max\_three\_objects                                & Return the three data with the maximum three values \{value\}.                                                                                                                                & len(obejcts)\textgreater{}3                                                                                                                                                             \\ \cline{2-4} 
                                      & min\_three\_objects                                & Return the three data with the minimum three values \{value\}.                                                                                                                                & len(obejcts)\textgreater{}3                                                                                                                                                             \\ \cline{2-4} 
                                      & second\_max\_object                                & Return the data with the second maximum value \{value\}.                                                                                                                                      & len(obejcts)\textgreater{}1                                                                                                                                                             \\ \cline{2-4} 
                                      & second\_min\_object                                & Return the data with the second minimum value \{value\}.                                                                                                                                      & len(obejcts)\textgreater{}1                                                                                                                                                             \\ \hline
value                                 & value\_of\_objects                                 & Return the values of data.                                                                                                                                                                    & -                                                                                                                                                                                       \\ \hline
\multirow{5}{*}{text\_information}    & color\_of\_objects                                 & Return the color of data.                                                                                                                                                                     & \begin{tabular}[c]{@{}l@{}}len(obejcts)=1, \\ chart type not in {[}heatmap, 3D-Bar, bubble{]}, \\ color\_group\_selection not in function chain\end{tabular}                         \\ \cline{2-4} 
                                      & groups\_of\_object                              & Return the groups of data.                                                                                                                                                                    & one\_object\_selection not in function chain                                                                                                                                            \\ \cline{2-4} 
                                      & legends\_of\_object                                & Return the legend of data.                                                                                                                                                                    & \begin{tabular}[c]{@{}l@{}}one\_object\_selection not in function chain,\\ legend\_selection not in function chain\end{tabular}                                                         \\ \cline{2-4} 
                                      & legend\_of\_one\_object\_value                     & Return the legend of the specific data with value \{value\}.                                                                                                                                  & len(obejcts)\textgreater{}1                                                                                                                                                             \\ \cline{2-4} 
                                      & group\_of\_one\_object\_value                   & Return the group of the specific data with value \{value\}.                                                                                                                                   & len(obejcts)\textgreater{}1                                                                                                                                                             \\ \hline
\multirow{3}{*}{if\_match\_condition} & if\_object\_that\_equal\_to\_value                 & Return if the data\textbackslash{}'s value is equal to \{value\}.                                                                                                                             & len(obejcts)=1                                                                                                                                                                          \\ \cline{2-4} 
                                      & if\_object\_that\_larger\_than\_value              & Return if the data\textbackslash{}'s value is larger/more than \{value\}.                                                                                                                     & len(obejcts)=1                                                                                                                                                                          \\ \cline{2-4} 
                                      & if\_object\_that\_smaller\_than\_value             & Return if the data\textbackslash{}'s value is smaller/less than \{value\}.                                                                                                                    & len(obejcts)=1                                                                                                                                                                          \\ \hline
\multirow{3}{*}{filter}               & objects\_that\_larger\_than\_value                 & Return data whose values are larger/more than \{value\}                                                                                                                                       & len(obejcts)\textgreater{}1                                                                                                                                                             \\ \cline{2-4} 
                                      & objects\_that\_smaller\_than\_value                & Return data whose value are smaller/less than \{value\}                                                                                                                                       & len(obejcts)\textgreater{}1                                                                                                                                                             \\ \cline{2-4} 
                                      & objects\_with\_same\_value                         & Return one group of data with the same value \{value\}.                                                                                                                                       & len(obejcts)\textgreater{}1                                                                                                                                                             \\ \hline
\multirow{4}{*}{count}                & count\_of\_objects                                 & Return the number of data, with values \{value\}.                                                                                                                                             & -                                                                                                                                                                                       \\ \cline{2-4} 
                                      & num\_of\_legends                                   & \begin{tabular}[c]{@{}l@{}}Return the number of legends used among the data, \\ with legends \{value\}.\end{tabular}                                                                          & -                                                                                                                                                                                       \\ \cline{2-4} 
                                      & num\_of\_colors                                    & \begin{tabular}[c]{@{}l@{}}Return the number of colors used among the data, \\ with colors \{value\}.\end{tabular}                                                                            & \begin{tabular}[c]{@{}l@{}}chart type not in  {[}heatmap, 3D-Bar, bubble{]},\\ color\_group\_selection not in function chain,\\ color\_selection not in function chain.\end{tabular} \\ \cline{2-4} 
                                      & num\_of\_groups                                 & \begin{tabular}[c]{@{}l@{}}Return the number of groups used among the data, \\ with group \{group name\}.\end{tabular}                                                                        & -                                                                                                                                                                                       \\ \hline
\multirow{2}{*}{exclude\_objects}     & exclude\_objects\_with\_groups                  & \begin{tabular}[c]{@{}l@{}}Exclude the data with the group \{group name\} \\ and return the data without the groups.\end{tabular}                                                             & group number\textgreater{}1                                                                                                                                                             \\ \cline{2-4} 
                                      & exclude\_objects\_with\_legends                    & \begin{tabular}[c]{@{}l@{}}Exclude the data with the legends \{legend name\} \\ and return the data without the legends.\end{tabular}                                                         & legend number\textgreater{}1                                                                                                                                                            \\ \hline
\multirow{2}{*}{min\_max\_diff\_arg}  & the\_group\_that\_has\_maximum\_difference      & \begin{tabular}[c]{@{}l@{}}Return the group B that has the maximum difference \\ between the two legends of data, \\ with value = max(|A1-A2|, |B1-B2|, |C1-C2|) = \{value\}.\end{tabular} & groun number \textgreater 1                                                                                                                                                             \\ \cline{2-4} 
                                      & the\_group\_that\_has\_minimum\_difference      & \begin{tabular}[c]{@{}l@{}}Return the group B that has the minimum difference \\ between the two legends of data, \\ with value = min(|A1-A2|, |B1-B2|, |C1-C2|) = \{value\}.\end{tabular} & groun number \textgreater 1                                                                                                                                                             \\ \hline
\multirow{6}{*}{if\_match\_condition} & if\_objects\_consistently\_increase                & Return if the values of the data consistently increase.                                                                                                                                       & \begin{tabular}[c]{@{}l@{}}legend\_selection or color\_selection in function chain, \\ chart type in {[}bar, line{]}.\end{tabular}                                                      \\ \cline{2-4} 
                                      & if\_objects\_consistently\_decrease                & Return if the values of the data consistently decrease.                                                                                                                                       & \begin{tabular}[c]{@{}l@{}}legend\_selection or color\_selection in function chain, \\ chart type in {[}bar, line{]}.\end{tabular}                                                      \\ \cline{2-4} 
                                      & if\_same\_values                                   & Return if the values of the data are the same.                                                                                                                                                & len(obejcts)\textgreater{}1                                                                                                                                                             \\ \cline{2-4} 
                                      & if\_same\_colors                                   & Return if the colors of the data are the same.                                                                                                                                                & len(obejcts)\textgreater{}1                                                                                                                                                             \\ \cline{2-4} 
                                      & if\_same\_groups                                & Return if the groups of the data are the same.                                                                                                                                                & len(obejcts)\textgreater{}1                                                                                                                                                             \\ \cline{2-4} 
                                      & if\_same\_legends                                  & Return if the legends of the data are the same.                                                                                                                                               & len(obejcts)\textgreater{}1                                                                                                                                                             \\ \hline
\multirow{18}{*}{position}            & upper\_one\_bar                                    & Return the upper-position bar in the chart.                                                                                                                                                   & chart type = bar                                                                                                                                                                        \\ \cline{2-4} 
                                      & upper\_two\_bars                                   & Return the upper two-position bars in the chart.                                                                                                                                              & chart type = bar                                                                                                                                                                        \\ \cline{2-4} 
                                      & upper\_three\_bars                                 & Return the upper three-position bars in the chart.                                                                                                                                            & chart type = bar                                                                                                                                                                        \\ \cline{2-4} 
                                      & bottom\_one\_bar                                   & Return the bottom bar in the chart.                                                                                                                                                           & chart type = bar                                                                                                                                                                        \\ \cline{2-4} 
                                      & bottom\_two\_bars                                  & Return the two bottom-position bars in the chart.                                                                                                                                             & chart type = bar                                                                                                                                                                        \\ \cline{2-4} 
                                      & bottom\_three\_bars                                & Return the three bottomposition bars in the chart.                                                                                                                                            & chart type = bar                                                                                                                                                                        \\ \cline{2-4} 
                                      & leftmost\_object                                   & Return the leftmost bars in the chart.                                                                                                                                                        & chart type in {[}bar, line{]}                                                                                                                                                           \\ \cline{2-4} 
                                      & left\_two\_objects                                 & Return the two leftmost bars in the chart.                                                                                                                                                    & chart type in {[}bar, line{]}                                                                                                                                                           \\ \cline{2-4} 
                                      & left\_three\_objects                               & Return the thr.ee leftmost bars in the chart                                                                                                                                                  & chart type in {[}bar, line{]}                                                                                                                                                           \\ \cline{2-4} 
                                      & rightmost\_object                                  & Return the rightmost bars in the chart.                                                                                                                                                       & chart type in {[}bar, line{]}                                                                                                                                                           \\ \cline{2-4} 
                                      & right\_two\_objects                                & Return the two rightmost bars in the chart.                                                                                                                                                   & chart type in {[}bar, line{]}                                                                                                                                                           \\ \cline{2-4} 
                                      & right\_three\_objects                              & Return the three rightmost bars in the chart.                                                                                                                                                 & chart type in {[}bar, line{]}                                                                                                                                                           \\ \cline{2-4} 
                                      & upper\_rightmost\_object                           & Return the upper-rightmost bar in the chart.                                                                                                                                                  & chart type = bar\_stacked                                                                                                                                                               \\ \cline{2-4} 
                                      & upper\_leftmost\_object                            & Return the upper-leftmost bar in the chart.                                                                                                                                                   & chart type = bar\_stacked                                                                                                                                                               \\ \cline{2-4} 
                                      & lower\_rightmost\_object                           & Return the bottom-rightmost bar in the chart.                                                                                                                                                 & chart type = bar\_stacked                                                                                                                                                               \\ \cline{2-4} 
                                      & lower\_leftmost\_object                            & Return the bottom-leftmost bar in the chart.                                                                                                                                                  & chart type = bar\_stacked                                                                                                                                                               \\ \cline{2-4} 
                                      & upper\_line\_of\_objects                           & Return the objects in the upper-position line of the chart.                                                                                                                                   & chart type = line                                                                                                                                                                       \\ \cline{2-4} 
                                      & lower\_line\_of\_objects                           & Return the objects in the bottom-position line of the chart.                                                                                                                                  & chart type = line                                                                                                                                                                       \\ \hline
\multirow{2}{*}{min\_max\_diff}       & maximum\_difference\_between\_two\_group\_of\_data & \begin{tabular}[c]{@{}l@{}}Return the maximum difference between the two legends of data:\\ max(|A1-A2|, |B1-B2|, |C1-C2|).\end{tabular}                                                      & groun number \textgreater 1                                                                                                                                                             \\ \cline{2-4} 
                                      & minimum\_difference\_between\_two\_group\_of\_data & \begin{tabular}[c]{@{}l@{}}Return the minimum difference between the two legends of data: \\ min(|A1-A2|, |B1-B2|, |C1-C2|).\end{tabular}                                                     & groun number \textgreater 1                                                                                                                                                             \\ \hline
\end{tabular}}
\caption{Overview of object functions.}
\vspace{-0.2in}
\label{object_function_details}
\end{table*}

\clearpage

\onecolumn

\section{Examples}
\label{examples}

\subsection{Example 1}
\label{example 4}
\begin{example}[Example]
\includegraphics[width=0.95\linewidth]{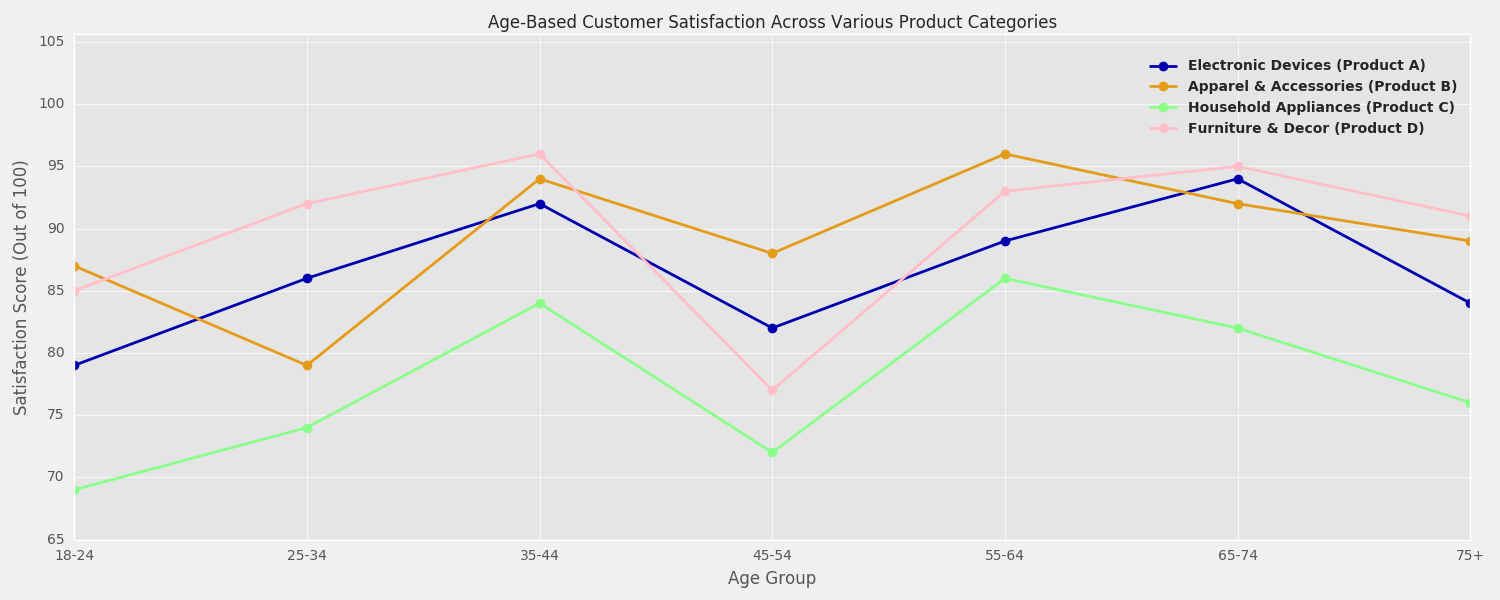}
\small{
\begin{lstlisting}[breaklines, basicstyle=\ttfamily]
Question: What is the satisfaction score for Apparel & Accessories (Product B) among customers in the age group 18-24 according to the chart?

Function chain: one_object_selection/value_of_objects

Answer: 87
\end{lstlisting}}
\end{example}

\begin{correctanswer}
\textbf{InternVL-2.5-8B}: \small{Rationale: The chart shows the satisfaction scores for different age groups and product categories. For Apparel \& Accessories (Product B), the satisfaction score for the age group 18-24 is indicated by the line that starts from the left side of the chart and rises to the right. The score for this age group is approximately 87 out of 100. Answer: 87}
\end{correctanswer}

\begin{correctanswer}
\textbf{Qwen2VL-7B}: \small{Rationale: The satisfaction score for Apparel \& Accessories (Product B) among customers in the age group 18-24 is found by looking at the orange line and the corresponding point on the x-axis for the 18-24 age group. The score is approximately 85.}
\end{correctanswer}

\begin{correctanswer}
\textbf{GPT4o}: \small{Rationale: To find the satisfaction score for Apparel \& Accessories (Product B) among customers in the age group 18-24, we need to look at the orange line on the chart, which represents Product B. At the age group 18-24, the orange line is at the satisfaction score of 85. Answer: 85}
\end{correctanswer}

\clearpage

\subsection{Example 2}
\label{example 1}
\begin{example}[Example]
\includegraphics[width=0.6\linewidth]{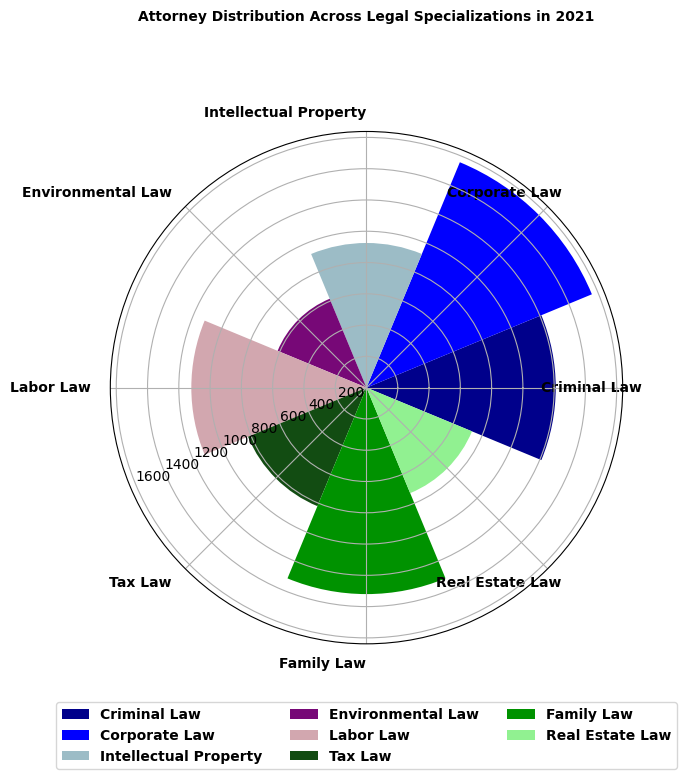}
\small{
\begin{lstlisting}[breaklines, basicstyle=\ttfamily]
Question: How many legal specializations have more than 925 attorneys according to the 'Attorney Distribution' group in the chart?

Function chain: group_selection/objects_that_larger_than_value/count_of_objects
Answer: 4
\end{lstlisting}}
\end{example}

\begin{wronganswer}
\textbf{InternVL-2.5-8B}: \small{Rationale: The chart shows the number of attorneys in each legal specialization. The specializations with more than 925 attorneys are Criminal Law, Corporate Law, and Intellectual Property. These three specializations have bars that extend beyond the 925 mark on the vertical axis. Answer: \red{3}.}
\end{wronganswer}

\begin{wronganswer}
\textbf{Qwen2VL-7B}: \small{Rationale: The 'Attorney Distribution' group in the chart shows that Corporate Law has more than 925 attorneys.}
\end{wronganswer}

\begin{wronganswer}
\textbf{GPT4o}: \small{Rationale: To determine how many legal specializations have more than 925 attorneys, we need to examine the chart and identify the specializations that extend beyond the 925 mark on the radial axis. The specializations and their approximate attorney counts are as follows:

1. Criminal Law: Approximately 1400 attorneys

2. Corporate Law: Approximately 1600 attorneys

3. Environmental Law: Approximately 1000 attorneys

4. Family Law: Approximately 1200 attorneys

5. Intellectual Property: Approximately 800 attorneys

6. Labor Law: Approximately 1100 attorneys

7. Real Estate Law: Approximately 1300 attorneys

8. Tax Law: Approximately 900 attorneys

The specializations with more than 925 attorneys are Criminal Law, Corporate Law, Environmental Law, Family Law, Labor Law, and Real Estate Law. Therefore, there are 6 specializations with more than 925 attorneys. Answer: \red{6}}
\end{wronganswer}

\clearpage

\subsection{Example 3}
\label{example 2}
\begin{example}[Example]
\includegraphics[width=0.8\linewidth]{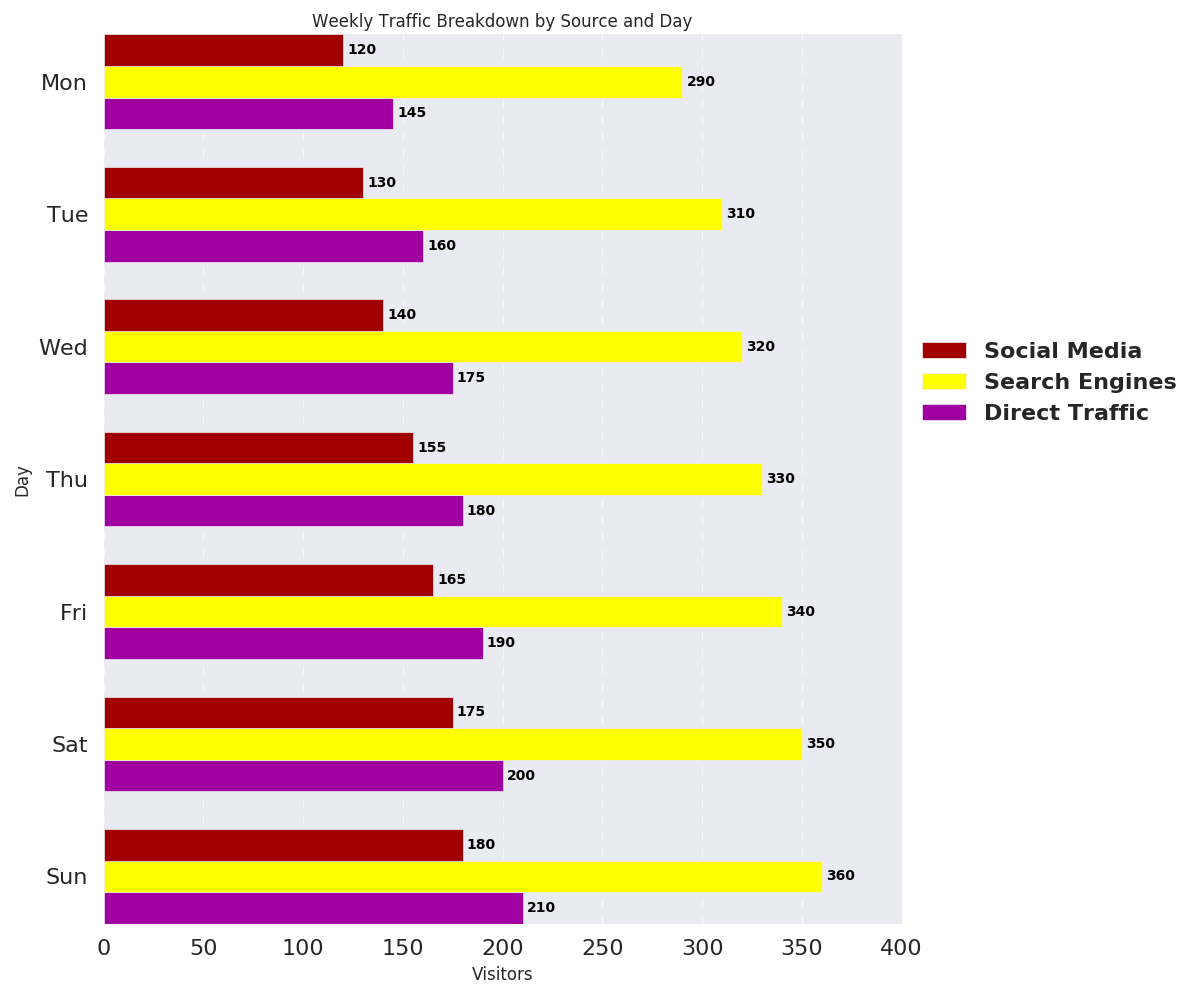}
\small{
\begin{lstlisting}[breaklines, basicstyle=\ttfamily]
Question: What is the ratio of Social Media traffic on Sunday compared to the day with the second highest Social Media traffic?

Function chain: one_object_selection/value_of_objects/legend_selection/second_max_object/value_of_objects/A_divided_by_B

Answer: 1.0285714285714285
\end{lstlisting}}
\end{example}

\begin{wronganswer}
\textbf{InternVL-2.5-8B}: \small{Rationale: The bar chart shows the weekly traffic breakdown by source and day. The day with the second highest Social Media traffic is Friday with \red{165} visitors. The ratio of Social Media traffic on Sunday compared to Friday is 180/165, which simplifies to 12/11 or approximately 1.09. Answer: \red{1.09}.}
\end{wronganswer}

\begin{wronganswer}
\textbf{Qwen2VL-7B}: \small{Rationale: To find the ratio of Social Media traffic on Sunday compared to the day with the second highest Social Media traffic, we need to identify the Social Media traffic for both days and then calculate the ratio. The Social Media traffic on Sunday is 180 visitors, and the day with the second highest Social Media traffic is Friday, with \red{165} visitors. The ratio is calculated as follows: 180 (Sunday) / 165 (Friday) = 1.103. Therefore, the ratio of Social Media traffic on Sunday compared to the day with the second highest Social Media traffic is approximately \red{1.103}.}
\end{wronganswer}

\begin{wronganswer}
\textbf{GPT4o}: \small{Rationale: From the chart, \red{the Social Media traffic for each day is as follows: Monday - 120, Tuesday - 130, Wednesday - 140, Thursday - 155, Friday - 165, Saturday - 175, Sunday - 210}. Sunday has the highest Social Media traffic with 210 visitors. The day with the second highest Social Media traffic is Saturday with 175 visitors. The ratio of Social Media traffic on Sunday compared to Saturday is 210:175. Simplifying this ratio by dividing both numbers by their greatest common divisor, which is 35, we get 6:5. Answer: \red{6:5}}
\end{wronganswer}

\clearpage

\subsection{Example 4}
\label{example 3}
\begin{example}[Example]
\includegraphics[width=0.8\linewidth]{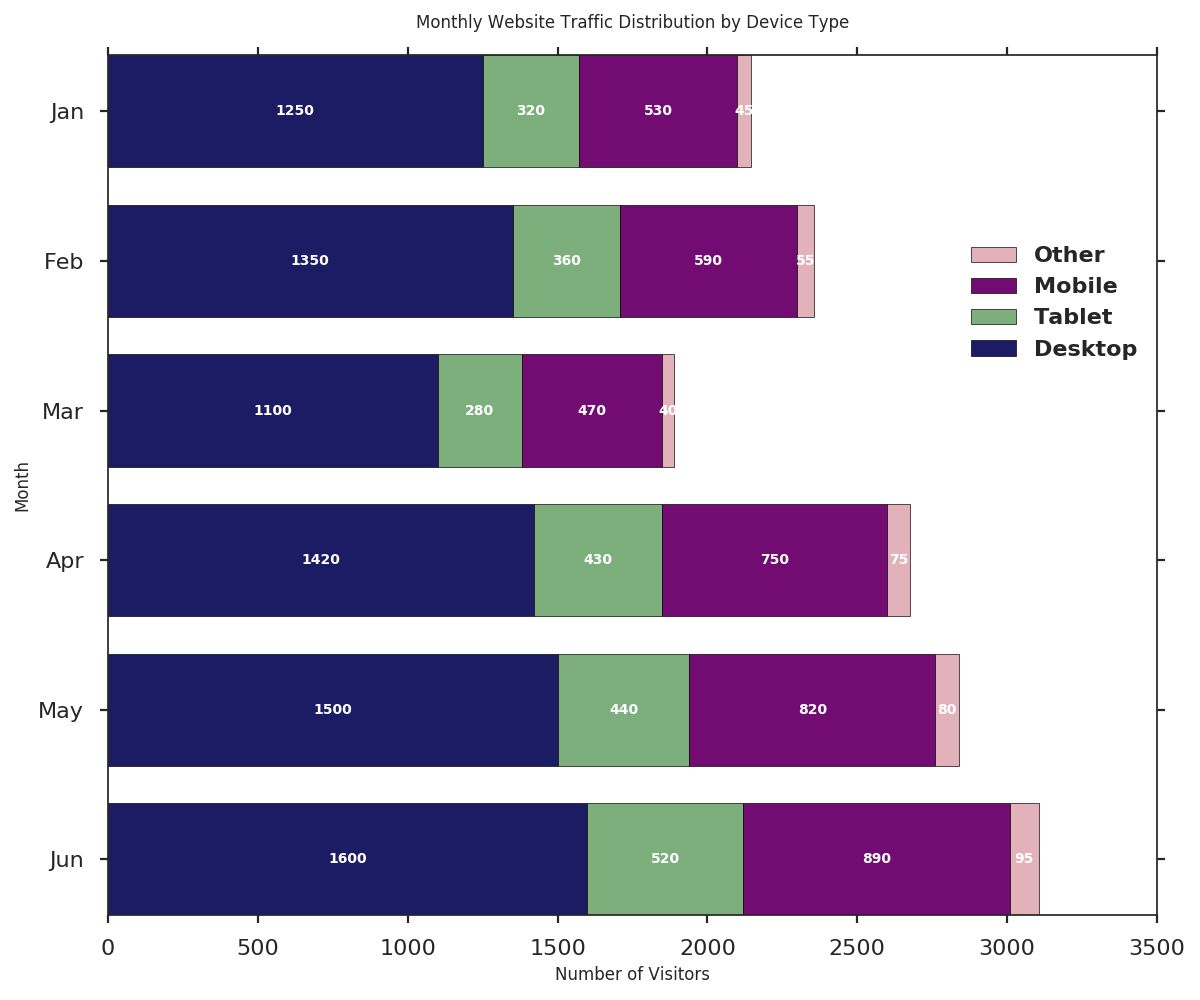}
\small{
\begin{lstlisting}[breaklines, basicstyle=\ttfamily]
Question: If we find the number of visitors using 'Other' devices in April, the number of 'Mobile' visitors in March, and the number of 'Tablet' visitors in May, what is the median value among these three numbers?

Function chain: one_object_selection/value_of_objects/one_object_selection/value_of_objects/one_object_selection/value_of_objects/median_of_values

Answer: 440
\end{lstlisting}}
\end{example}

\begin{correctanswer}
\textbf{InternVL-2.5-8B}: \small{Rationale: The number of visitors using 'Other' devices in April is 75, the number of 'Mobile' visitors in March is 470, and the number of 'Tablet' visitors in May is 440. To find the median, we arrange these numbers in ascending order: 75, 440, 470. The median is the middle number, which is 440. Answer: 440}
\end{correctanswer}

\begin{wronganswer}
\textbf{Qwen2VL-7B}: \small{Rationale: To find the median value among the three numbers, we first need to arrange the numbers in ascending order. The numbers are: 75 (April), 80 (May), and 40 (March). After arranging them, the middle number is the median. In this case, the median is 75. Answer: \red{75}.}
\end{wronganswer}

\begin{correctanswer}
\textbf{GPT4o}: \small{Rationale: To find the median value among the three specified numbers, we first need to identify each number from the chart. The number of 'Other' device visitors in April is 75. The number of 'Mobile' visitors in March is 470. The number of 'Tablet' visitors in May is 440. Now, we have the three numbers: 75, 470, and 440. To find the median, we arrange these numbers in ascending order: 75, 440, 470. The median is the middle number in this ordered list, which is 440. Answer: 440}
\end{correctanswer}

\clearpage

\subsection{Example 5}
\label{example 5}
\begin{example}[Example]
\includegraphics[width=0.8\linewidth]{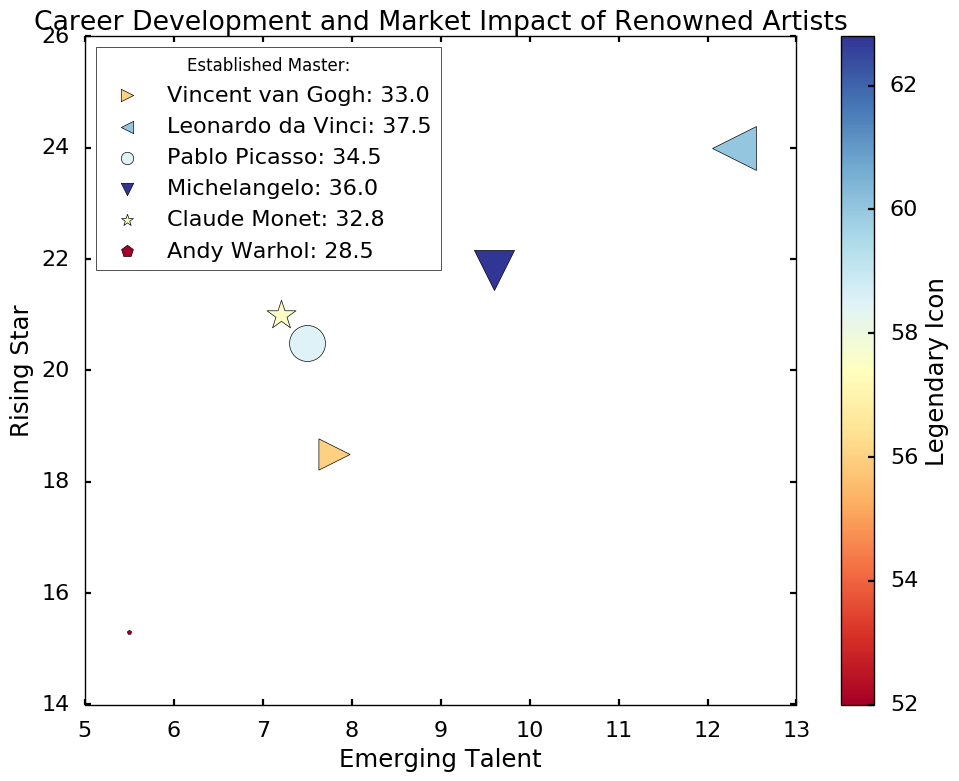}
\small{
\begin{lstlisting}[breaklines, basicstyle=\ttfamily]
Question: What is the minimum market value (in millions) shown for any artist during their respective phase in the chart, considering the career development and market impact of renowned artists categorized into Emerging Talent, Rising Star, Established Master, and Legendary Icon?

Function chain: all_object_selection/min_one_object/value_of_objects

Answer: 5.5
\end{lstlisting}}
\end{example}

\begin{wronganswer}
\textbf{InternVL-2.5-8B}: \small{Rationale: The chart shows different artists at various stages of their career development and market impact. The minimum value is represented by the artist with the lowest market value in the \"Emerging Talent\" phase, which is Andy Warhol with a value of 28.5 million. Answer: \red{28.5}}
\end{wronganswer}

\begin{correctanswer}
\textbf{Qwen2VL-7B}: \small{Rationale: The minimum market value shown for any artist during their respective phase is 5.5 million, which is represented by the smallest data point in the chart.}
\end{correctanswer}

\begin{wronganswer}
\textbf{GPT4o}: \small{Rationale: The chart shows the market values of various renowned artists during their respective phases. The artists listed are Vincent van Gogh, Leonardo da Vinci, Pablo Picasso, Michelangelo, Claude Monet, and Andy Warhol. The market values are given in millions: Vincent van Gogh (33.0), Leonardo da Vinci (37.5), Pablo Picasso (34.5), Michelangelo (36.0), Claude Monet (32.8), and Andy Warhol (28.5). Among these values, Andy Warhol has the lowest market value at 28.5 million. Answer: 28.5}
\end{wronganswer}

\clearpage

\subsection{Example 6}
\label{example 6}
\begin{example}[Example]
\includegraphics[width=0.95\linewidth]{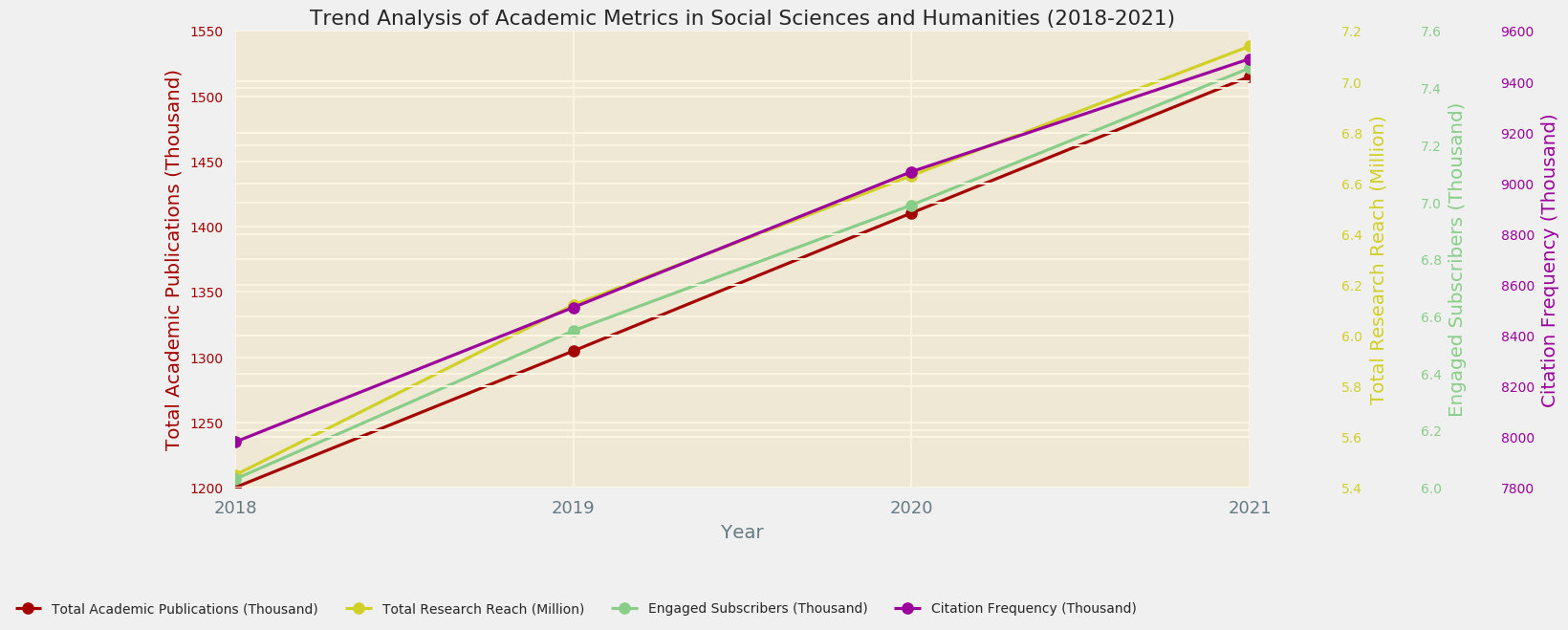}
\small{
\begin{lstlisting}[breaklines, basicstyle=\ttfamily]
Question: What was the total number of academic publications (in thousands) in the year 2018 according to the chart?

Function chain: one_object_selection/value_of_objects

Answer: 1200.2
\end{lstlisting}}
\end{example}

\begin{correctanswer}
\textbf{InternVL-2.5-8B}: \small{Rationale: The chart shows the total number of academic publications in thousands for the years 2018 to 2021. In 2018, the line representing total academic publications starts at the point corresponding to 1200 thousand. This indicates that the total number of academic publications in the year 2018 was 1200 thousand. Answer: 1200.}
\end{correctanswer}

\begin{correctanswer}
\textbf{Qwen2VL-7B}: \small{Rationale: The total number of academic publications in 2018 is represented by the red line on the chart. The value is approximately 1200 thousand.}
\end{correctanswer}

\begin{correctanswer}
\textbf{GPT4o}: \small{Rationale: The chart shows the trend of various academic metrics from 2018 to 2021. The red line represents the total academic publications in thousands. In 2018, the red line is at the 1250 mark on the left y-axis, which indicates the total number of academic publications in thousands for that year. Answer: 1250}
\end{correctanswer}

\clearpage

\subsection{Example 7}
\label{example 7}
\begin{example}[Example]
\includegraphics[width=0.95\linewidth]{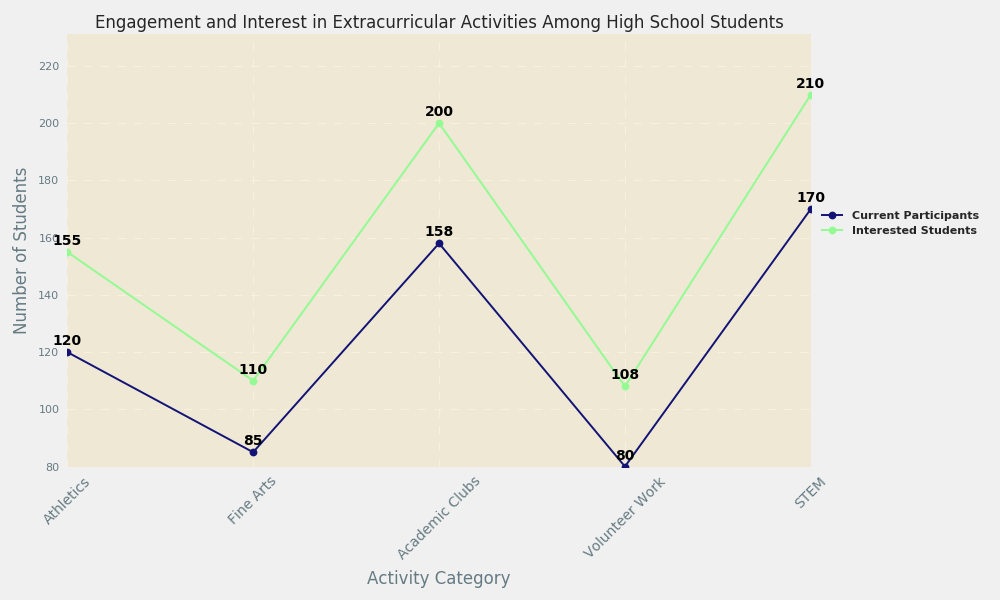}
\small{
\begin{lstlisting}[breaklines, basicstyle=\ttfamily]
Question: If we find the number of current participants in Volunteer Work, identify the maximum number of interested students across all activity categories, and also find the minimum number of current participants across all activity categories, what is the sum of these three values?

Function chain: one_object_selection/value_of_objects/legend_selection/max_one_object/value_of_objects/legend_selection/min_one_object/value_of_objects/sum_of_values

Answer: 370
\end{lstlisting}}
\end{example}

\begin{correctanswer}
\textbf{InternVL-2.5-8B}: \small{Rationale: The number of current participants in Volunteer Work is 80. The maximum number of interested students across all activity categories is 210 (in STEM). The minimum number of current participants across all activity categories is 80 (in Volunteer Work). The sum of these three values is 80 + 210 + 80 = 370. Answer: 370}
\end{correctanswer}

\begin{wronganswer}
\textbf{Qwen2VL-7B}: \small{Rationale: To find the sum of the three values, we need to identify the number of current participants in Volunteer Work, the maximum number of interested students across all activity categories, and the minimum number of current participants across all activity categories. Then, we add these three values together. Answer: 468}
\end{wronganswer}

\begin{correctanswer}
\textbf{GPT4o}: \small{Rationale: The number of current participants in Volunteer Work is 80. The maximum number of interested students across all activity categories is 210 (in STEM). The minimum number of current participants across all activity categories is 80 (in Volunteer Work). The sum of these three values is 80 + 210 + 80 = 370. Answer: 370}
\end{correctanswer}

\clearpage

\subsection{Example 8}
\label{example 8}
\begin{example}[Example]
\includegraphics[width=0.95\linewidth]{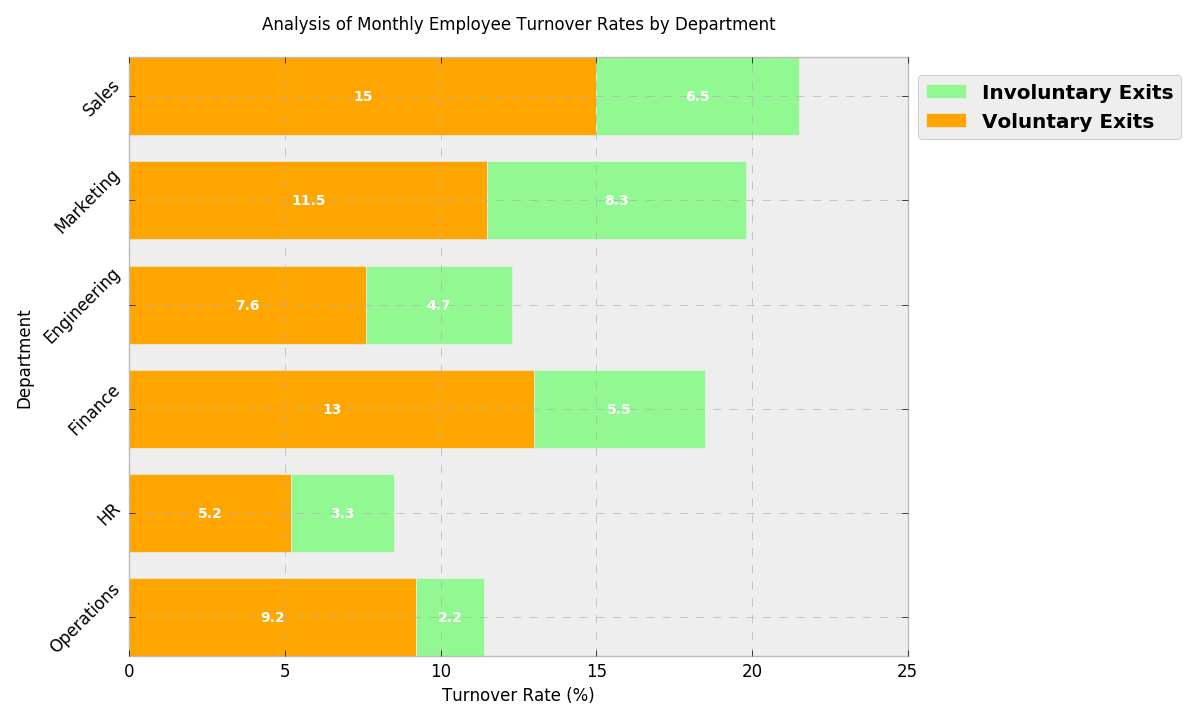}
\small{
\begin{lstlisting}[breaklines, basicstyle=\ttfamily]
Question: If we get the highest 'Involuntary Exits' rate among all departments and add it to the 'Involuntary Exits' rate of 'Sales', and then add the 'Voluntary Exits' rate of 'Operations', what is the total rate?

Function chain: legend_selection/max_one_object/value_of_objects/one_object_selection/value_of_objects/color_group_selection/value_of_objects/sum_of_values

Answer: 24.0%
\end{lstlisting}}
\end{example}

\begin{correctanswer}
\textbf{InternVL-2.5-8B}: \small{Rationale: The highest 'Involuntary Exits' rate among all departments is 8.3\% (Marketing). Adding this to the 'Involuntary Exits' rate of 'Sales' (6.5\%) gives us 14.8\%. The 'Voluntary Exits' rate of 'Operations' is 9.2\%. Adding this to the previous sum (14.8\%) gives us a total rate of 24\%. Answer: 24\%}
\end{correctanswer}

\begin{wronganswer}
\textbf{Qwen2VL-7B}: \small{Rationale: To find the total rate, we need to add the highest 'Involuntary Exits' rate, which is 6.5 from 'Sales', to the 'Involuntary Exits' rate of 'Sales', which is 6.5, and then add the 'Voluntary Exits' rate of 'Operations', which is 9.2. So, the total rate is 6.5 (from 'Sales') + 6.5 (from 'Sales') + 9.2 (from 'Operations') = 22.2.}
\end{wronganswer}

\begin{correctanswer}
\textbf{GPT4o}: \small{Rationale: First, identify the highest 'Involuntary Exits' rate among all departments. From the chart, the 'Involuntary Exits' rates are: Sales (6.5), Marketing (8.3), Engineering (4.7), Finance (5.5), HR (3.3), and Operations (2.2). The highest rate is 8.3 from Marketing. Next, add this to the 'Involuntary Exits' rate of 'Sales', which is 6.5. So, 8.3 + 6.5 = 14.8. Finally, add the 'Voluntary Exits' rate of 'Operations', which is 9.2. Therefore, 14.8 + 9.2 = 24.0. Answer: 24.0}
\end{correctanswer}

\clearpage

\subsection{Example 9}
\label{example 9}
\begin{example}[Example]
\includegraphics[width=0.8\linewidth]{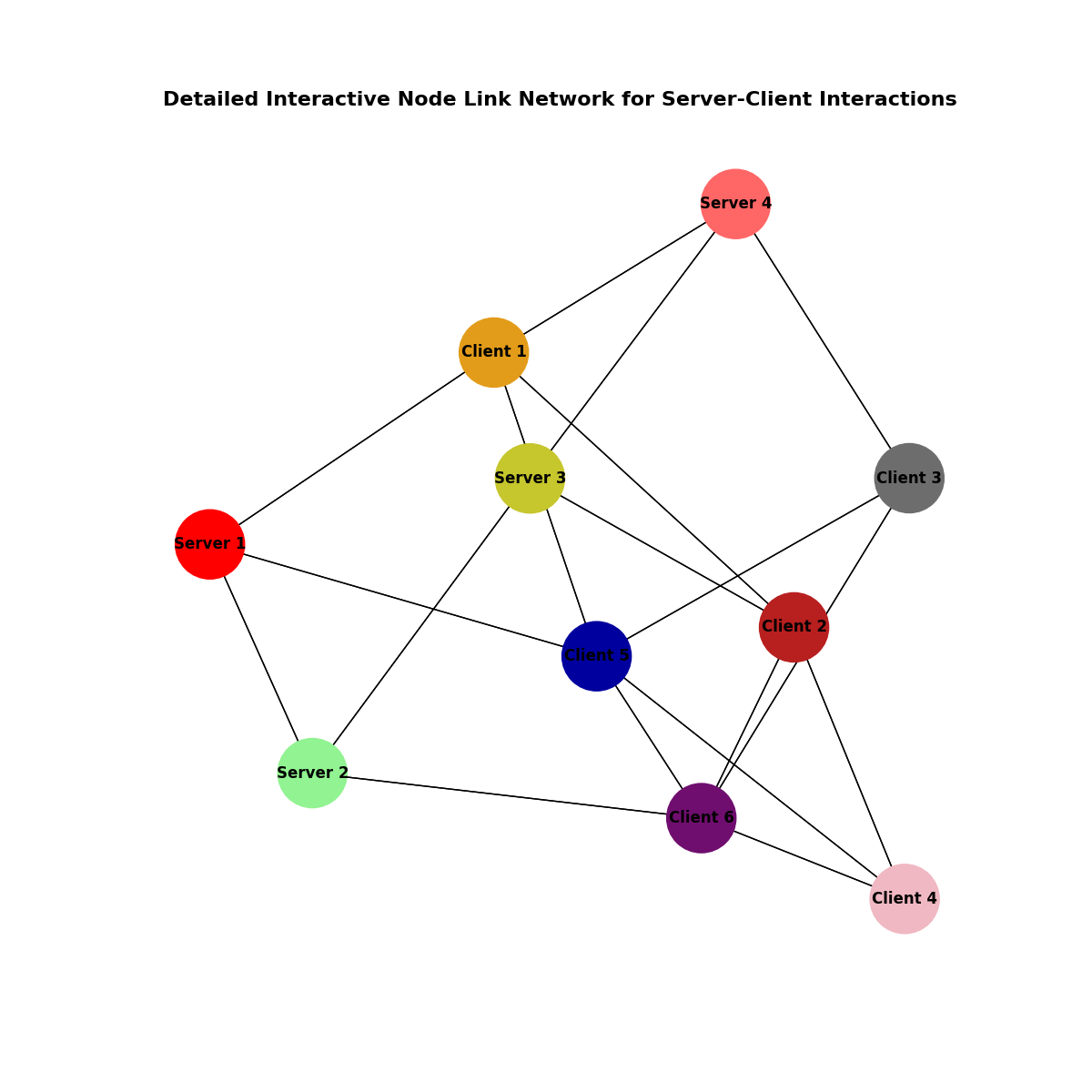}
\small{
\begin{lstlisting}[breaklines, basicstyle=\ttfamily]
Question: In the Detailed Interactive Node Link Network for Server-Client Interactions, is the node represented by 'Client 5' connected to the node represented by 'Client 3'?

Function chain: one_object_selection/if_object_connect_to_A

Answer: Yes
\end{lstlisting}}
\end{example}

\begin{wronganswer}
\textbf{InternVL-2.5-8B}: \small{Rationale: The diagram shows a network of connections between servers and clients. 'Client 5' is connected to 'Server 3' and 'Server 2', while 'Client 3' is connected to 'Server 4'. There is no direct connection between 'Client 5' and 'Client 3' in the diagram. Answer: no}
\end{wronganswer}

\begin{correctanswer}
\textbf{Qwen2VL-7B}: \small{Rationale: The node represented by 'Client 5' is connected to the node represented by 'Client 3' as indicated by the line connecting them in the network diagram. Answer: Yes}
\end{correctanswer}

\begin{wronganswer}
\textbf{GPT4o}: \small{Rationale: To determine if 'Client 5' is connected to 'Client 3', we need to trace the lines (edges) from 'Client 5' to see if there is a direct connection to 'Client 3'. Upon examining the network, 'Client 5' is connected to 'Server 3', 'Client 2', and 'Client 6'. 'Client 3' is connected to 'Server 4', 'Server 3', and 'Client 2'. There is no direct line connecting 'Client 5' to 'Client 3'. Therefore, 'Client 5' is not directly connected to 'Client 3'. Answer: No}
\end{wronganswer}

\end{document}